\def\year{2020}\relax
\documentclass[letterpaper]{article} %
\usepackage{aaai20}  %
\usepackage{times}  %
\usepackage{helvet} %
\usepackage{courier}  %
\usepackage[hyphens]{url}  %
\usepackage{graphicx} %
\usepackage{subcaption}
\usepackage[normalem]{ulem}
\usepackage{algorithm}%
\usepackage{algpseudocode}%
\usepackage{amsmath}
\usepackage{mathrsfs}
\usepackage{amsfonts}
\usepackage{amsthm}

\usepackage{booktabs}
\usepackage{graphicx}  %
\newtheorem{defn}{Definition}
\newcommand{\indic}[1]{\mathbb{I}_{\left [ #1 \right ]}}
\newcommand{\cbar}{\, | \,}
\DeclareMathOperator*{\expect}{{\huge \mathbb{E}}}

\newcommand{\citet}[1]{\citeauthor{#1} \shortcite{#1}}
\newcommand{\citep}[1]{(\citeauthor{#1} \citeyear{#1})}

\makeatletter
\newcommand{\algmargin}{\the\ALG@thistlm}   
\makeatother
\algnewcommand{\parState}[1]{\State%
    \parbox[t]{\dimexpr\linewidth-\algmargin}{\strut #1\strut}}

\title{Algorithmic Improvements for Deep Reinforcement Learning applied to Interactive Fiction}
\author{
Vishal Jain,\textsuperscript{\rm 1, \rm 3}
William Fedus,\textsuperscript{\rm 1, \rm 2}
Hugo Larochelle,\textsuperscript{\rm 1, \rm 2, \rm 5}
Doina Precup,\textsuperscript{\rm 1, \rm 3, \rm 4, \rm 5 }
Marc G. Bellemare\textsuperscript{\rm 1, \rm 2, \rm 3, \rm 5}\\
\textsuperscript{\rm 1}Mila,
\textsuperscript{\rm 2}Google Brain,
\textsuperscript{\rm 3}McGill University,
\textsuperscript{\rm 4}DeepMind,
\textsuperscript{\rm 5}CIFAR Fellow\\
vishal.jain@mail.mcgill.ca, liamfedus@google.com, hugolarochelle@google.com.
dprecup@cs.mcgill.ca, bellemare@google.com
}

\newcommand{\cA}{\mathcal{A}}
\newcommand{\hatxi}{\hat \xi}
\newcommand{\hatcAt}{\hat \cA_t}
\newcommand{\eqnref}[1]{(\ref{eqn:#1})}

\begin{document}

\maketitle

\begin{abstract}
Text-based games are a natural challenge domain for deep reinforcement learning algorithms. Their state and action spaces are combinatorially large, their reward function is sparse, and they are partially observable: the agent is informed of the consequences of its actions through textual feedback. In this paper we emphasize this latter point and consider the design of a deep reinforcement learning agent that can play from feedback alone. Our design recognizes and takes advantage of the structural characteristics of text-based games. We first propose a contextualisation mechanism, based on accumulated reward, which simplifies the learning problem and mitigates partial observability. We then study different methods that rely on the notion that most actions are ineffectual in any given situation, following Zahavy et al.'s idea of an admissible action. We evaluate these techniques in a series of text-based games of increasing difficulty based on the TextWorld framework, as well as the iconic game \textsc{Zork}. Empirically, we find that these techniques improve the performance of a baseline deep reinforcement learning agent applied to text-based games.
\end{abstract}

\section{Introduction}
\label{introduction}
\begin{figure}[ht]
\includegraphics[scale=0.2]{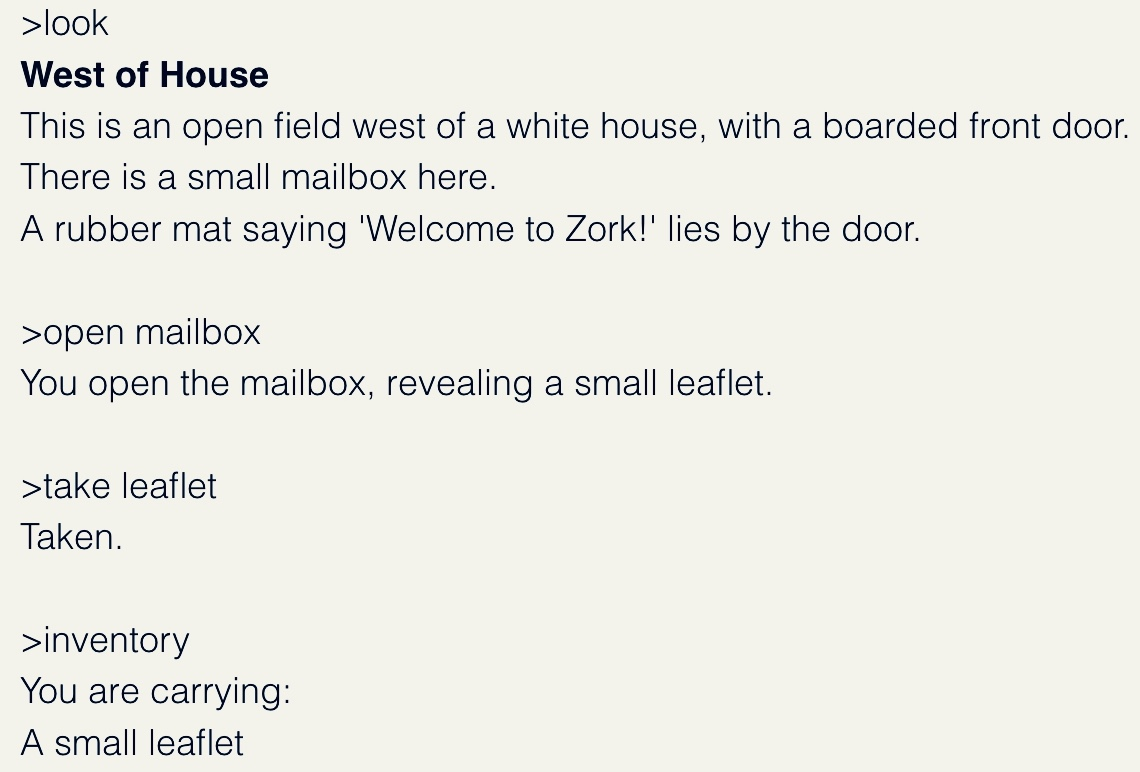}
\caption{The introductory gameplay from \textsc{Zork}.}
\label{fig:gameplay}
\end{figure}

In a text-based game, also called \emph{interactive fiction} (IF), an agent interacts with its environment through a natural language interface. Actions consist of short textual commands, while observations are paragraphs describing the outcome of these actions (Figure \ref{fig:gameplay}). Recently, interactive fiction has emerged as an important challenge for AI techniques \citep{atkinson2018textbased}, in great part because the genre combines natural language with sequential decision-making.

From a reinforcement learning perspective, IF domains pose a number of challenges. First, the state space is typically combinatorial in nature, due to the presence of objects and characters with which the player can interact. Since any natural language sentence may be given as a valid command, the action space is similarly combinatorial. 
The player observes its environment through feedback in natural language, making this a partially observable problem. The reward structure is usually sparse, with non-zero rewards only received when the agent accomplishes something meaningful, such as retrieving an important object or unlocking a new part of the domain.

We are particularly interested in bringing deep reinforcement learning techniques to bear on this problem. In this paper, we consider how to design an agent architecture that can learn to play text adventure games from feedback alone. Despite the inherent challenges of the domain, we identify three structural aspects that make progress possible:
\begin{itemize}
    \item \textbf{Rewards from subtasks.} The optimal behaviour completes a series of subtasks towards the eventual game end;
    \item \textbf{Transition structure.} Most actions have no effect in a given state;
    \item \textbf{Memory as state.} Remembering key past events is often sufficient to deal with partial observability.
\end{itemize}
While these properties have been remarked on in previous work \cite{Narasimhan_2015,Zahavy:2018:LLA:3327144.3327274}, here
we relax some of the assumptions previously made and provide fresh tools to more tractably solve IF domains. More generally, we believe these tools to be useful in partially observable domains with similar structure.

Our first contribution takes advantage of the special reward structure of IF domains. In IF, the accumulated reward within an episode correlates with the number of completed subtasks and provides a good proxy for an agent's progress. Our \emph{score contextualisation} architecture makes use of this fact by defining a piecewise value function composed of different deep network heads, where each piece corresponds to a particular level of cumulative reward. This separation allows the network to learn separate value functions for different portions of the complete task; in particular, when the problem is linear (i.e., there is a fixed ordering in which subtasks must be completed), our method can be used to learn a separate value function for each subtask.

Our second contribution extends the work of \citet{Zahavy:2018:LLA:3327144.3327274} on action elimination. We make exploration and action selection more tractable by determining which actions are \emph{admissible} in the current state. Formally, we say that an action is admissible if it leads to a change in the underlying game state. While the set of available actions is typically large in IF domains, there are usually few commands that are actually admissible in any particular context. Since the state is not directly observable, we first learn an LSTM-based \emph{auxiliary classifier} that predicts which actions are admissible given the agent's history of recent feedback. We use the predicted probability of an action being admissible to modulate or \emph{gate} which actions are available to the agent at each time step. We propose and compare three simple modulation methods: masking, drop out, and finally consistent Q-learning \citep{Bellemare:2016:IAG:3016100.3016105}. Compared to \citeauthor{Zahavy:2018:LLA:3327144.3327274}'s algorithm, our techniques are simpler in spirit and can be learned from feedback alone.

We show the effectiveness of our methods on a suite of seven IF problems of increasing difficulty generated using the TextWorld platform \citep{cote18textworld}. We find that combining the score contextualisation approach to an otherwise standard recurrent deep RL architecture leads to faster learning than when using a single value function. Furthermore, our action gating mechanism enables the learning agent to progress on the harder levels of our suite of problems.

\section{Problem Setting}
We represent an interactive fiction environment as a partially observable Markov decision process (POMDP) with deterministic observations. This POMDP is summarized by the tuple $(\mathcal{S},\mathcal{A},P,r,\mathcal{O},\psi, \gamma)$, where $\mathcal{S}$ is the state space, $\mathcal{A}$ the action space, $P$ is the transition function, $r:\mathcal{S}\times\mathcal{A}\to\mathbb R$ is the reward function, and $\gamma \in [0,1)$ is the discount factor. The function $\psi:\mathcal{S}\times\mathcal{A}\times\mathcal{S}\to\mathcal{O}$
describes the observation $o = \psi(s,a,s')$ provided to the agent when action $a$ is taken in state $s$ and leads to state $s'$.

Throughout we will make use of standard notions from reinforcement learning \citep{sutton98reinforcement} as adapted to the POMDP literature \cite{mccallum95reinforcement,silver10montecarlo}. At time step $t$, the agent selects an action according to a policy $\pi$ which maps a \emph{history} $h_t := o_1, a_1, \dots, o_t$ to a distribution over actions, denoted $\pi(\cdot \cbar h_t)$. 
This history is a sequence of observations and actions which, from the agent's perspective, replaces the unobserved environment state $s_t$.
We denote by $B(s \cbar h_t)$ the probability or \emph{belief} of being in state $s$ after observing $h_t$. Finally, we will find it convenient to rely on time indices to indicate the relationship between a history $h_t$ and its successor, and denote by $h_{t+1}$ the history resulting from taking action $a_t$ in $h_t$ and observing $o_{t+1}$ as emitted by the hidden state $s_{t+1}$.

The action-value function $Q^\pi$ describes the expected discounted sum of rewards when choosing action $a$ after observing history $h_t$, and subsequently following policy $\pi$:
\begin{equation*}
    Q^\pi(h_t, a) = \expect \big [ \sum_{i \ge 0} \gamma^i r(s_{t+i}, a_{t+i}) \cbar h_t, a \big ],
\end{equation*}
where we assume that the action at time $t+j$ is drawn from $\pi(\cdot \, | \, h_{t+j})$; note that the reward depends on the sequence  of hidden states $s_{t+1}, s_{t+2}, \dots$ implied by the belief state $B(\cdot \cbar h_t)$.
The action-value function satisfies the Bellman equation over histories
\begin{equation*}
    Q^\pi(h_t, a) = \expect_{s_t, s_{t+1}} \big [ r(s_t, a) + \gamma \max_{a' \in \cA} Q^\pi(h_{t+1}, a') \big ] .
\end{equation*}
When the state is observed at each step ($\mathcal{O} = \mathcal{S}$), this simplifies to the usual Bellman equation for Markov decision processes:
\begin{equation}
    Q^\pi(s_t, a) = r(s_t, a) + \gamma \expect_{s_{t+1} \sim P} \max_{a' \in \cA} Q^\pi(s_{t+1}, a') . \label{eqn:bellman_equation}
\end{equation}
In the fully observable case we will conflate $s_t$ and $h_t$.

The Q-learning algorithm \citep{watkins89learning} over histories maintains an approximate action-value function $Q$ which is updated from samples $h_t, a_t, r_t, o_{t+1}$ using a step-size parameter $\alpha \in [0, 1)$:
\begin{align}
    Q(h_t, a_t) &\gets Q(h_t, a_t) + \alpha \delta_t \nonumber \\
    \delta_t &= r_t + \gamma \max_{a \in \mathcal{A}} Q(h_{t+1}, a) - Q(h_t, a_t) . \label{eqn:q_learning}
\end{align}
Q-learning is used to estimate the \emph{optimal action-value function}  attained by a policy which maximizes $Q^\pi$ for all histories. In the context of our work, we will assume that this policy exists.
Storing this action-value function in a lookup table is impractical, as there are in general an exponential number of histories to consider. Instead, we use recurrent neural networks approximate the Q-learning process.

\subsection{Consistent Q-Learning}

Consistent Q-learning \citep{Bellemare:2016:IAG:3016100.3016105} learns a value function which is consistent with respect to a local form of policy stationarity. Defined for a Markov decision process, it replaces the term $\delta_t$ in \eqnref{q_learning} by
\begin{small}
\begin{equation}
\delta^{\textsc{cql}}_t = r_t + \left \{ \begin{array}{cc} 
        \gamma \max_{a \in \cA} Q(s_{t+1}, a) - Q(s_t, a_t) & s_{t+1} \ne s_t \\
        (\gamma - 1) Q(s_t, a_t) & s_{t+1} = s_t .
    \end{array} \right . \label{eqn:consistent_bellman_operator}
\end{equation}
\end{small}
Consistent Q-learning can be shown to decrease the action-value of suboptimal actions while maintaining the action-value of the optimal action, leading to larger \emph{action gaps} and a potentially easier value estimation problem.

Observe that consistent Q-learning is not immediately adaptable to the history-based formulation, since $h_{t+1}$ and $h_t$ are sequences of different lengths (and therefore not comparable). One of our contributions in this paper is to derive a related algorithm suited to the history-based setting.

\subsection{Admissible Actions}

We will make use of the notion of an \emph{admissible action}, following terminology by \citet{Zahavy:2018:LLA:3327144.3327274}.\footnote{Note that our definition technically differs from \citet{Zahavy:2018:LLA:3327144.3327274}'s, who define an admissible action as one that is not ruled out by the learning algorithm.}
\begin{defn}
An action $a$ is \emph{admissible in state $s$} if
\begin{equation*}
    P(s \cbar s, a) < 1 .
\end{equation*}
That is, $a$ is admissible in $s$ if its application \emph{may} result in a change in the environment state. When $P(s \cbar s, a) = 1$, we say that an action is \emph{inadmissible}.
\end{defn}
We extend the notion of admissibility to histories as follows. We say that an action $a$ is admissible given a history $h$ if it is admissible in \emph{some} state that is possible given $h$, or equivalently:
\begin{equation*}
    \sum_{s \in \mathcal{S}} B(s \cbar h) P(s \cbar s, a) < 1.
\end{equation*}
We denote by $\xi(s) \subseteq \mathcal{A}$ the set of admissible actions in state $s$. Abusing notation, we define the \emph{admissibility function}
\begin{align*}
    \xi(s, a) &:= \indic{a \in \xi(s)} \\ \xi(h, a) &:= \Pr \{ a \in \xi(S) \}, S \sim B(\cdot \cbar h) .
\end{align*}
We write $\cA_t$ for the set of admissible actions given history $h_t$, i.e. the actions whose admissibility in $h_t$ is strictly greater than zero.
In IF domains, inadmissible actions are usually dominated, and we will deprioritize or altogether rule them out based on our estimate of $\xi(h, a)$.

\section{More Efficient Learning for IF Domains}
\label{methodology}
\begin{figure}[ht]
\centering
\includegraphics[width=\linewidth]{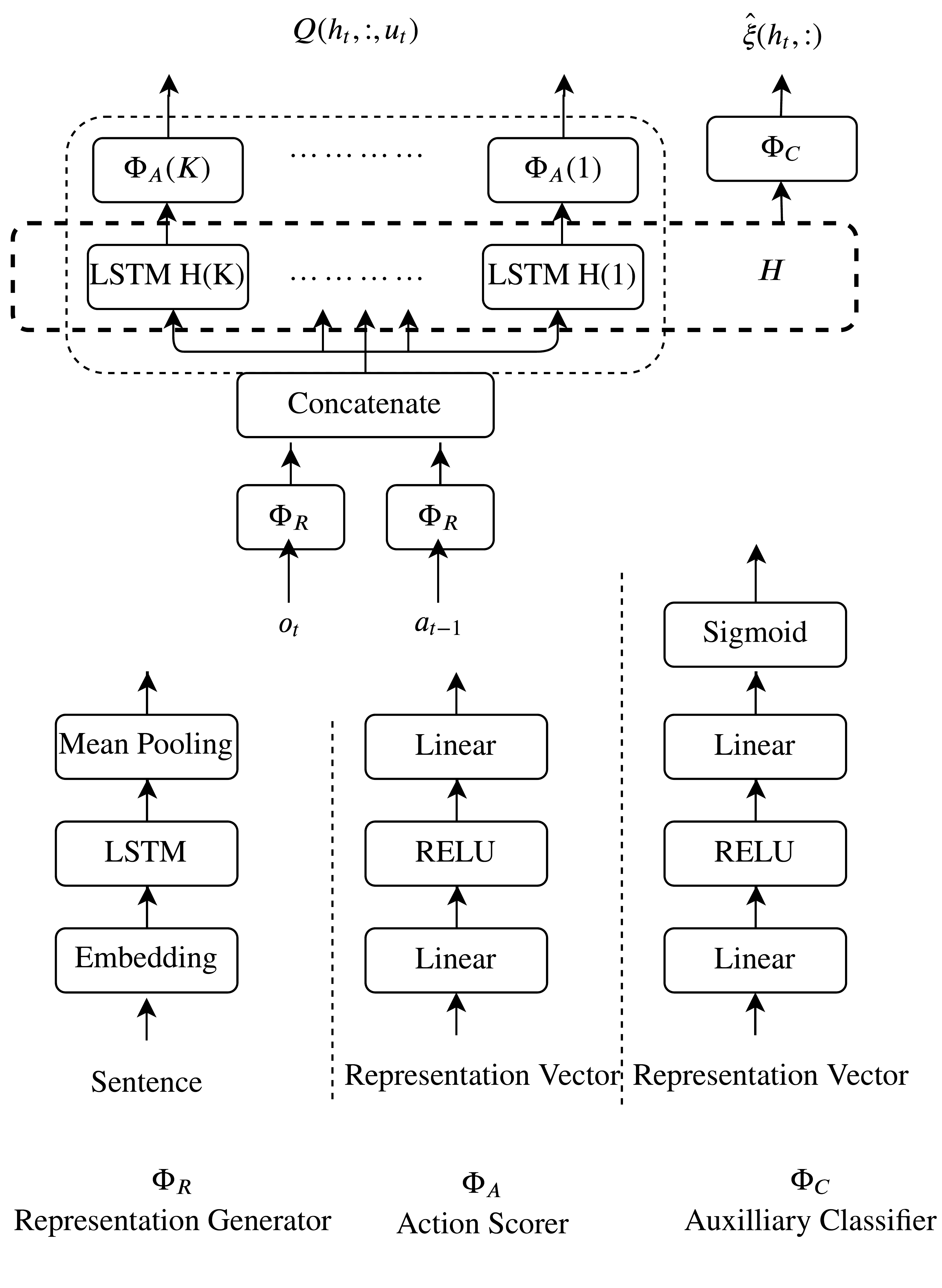}
\caption{Our IF architecture consists of three modules: a representation generator $\Phi_R$ that learns an embedding for a sentence, an action scorer $\Phi_A$ that chooses a network head $i$ (a feed-forward network) conditional on score $u_t$, learns its Q-values and outputs $Q(h_t,:,u_t)$ and finally, an auxilliary classifier $\Phi_C$ that learns an approximate admissibility function $\hatxi(h_t, :)$. The architecture is trained end-to-end.}
\label{fig:memento_architecture}
\end{figure}

We are interested in learning an action-value function which is close to optimal and from which can be derived a near-optimal policy. We would also like learning to proceed in a sample-efficient manner. In the context of IF domains, this is hindered by both the partially observable nature of the environment and the size of the action space. In this paper we propose two complementary ideas that alleviate some of the issues caused by partial observability and large action sets. The first idea contextualizes the action-value function on a surrogate notion of progress based on total reward so far, while the second seeks to eliminate inadmissible actions from the exploration and learning process.

Although our ideas are broadly applicable, for concreteness we describe their implementation in a deep reinforcement learning framework. Our agent architecture (Figure \ref{fig:memento_architecture}) is derived from the LSTM-DRQN agent \citep{yuan2018counting} and the work of \citet{Narasimhan_2015}.

\subsection{Score Contextualisation}

In applying reinforcement learning to games, it is by now customary to translate the player's score differential into rewards \cite{Bellemare_2013,openai2018}. Our setting is similar to Arcade Learning Environment in the sense that the environment provides the score. In IF, the player is awarded points for acquiring an important object, or completing some task relevant to progressing through the game. These awards occur in a linear, or almost linear structure, reflecting the agent's progression through the story, and are relatively sparse.
We emphasize that this is in contrast to the more general reinforcement learning setting, which may provide reward for surviving, or achieving something at a certain rate. In the video game \textsc{Space Invaders}, for example, the notion of ``finishing the game'' is ill-defined: the player's objective is to keep increasing their score until they run out of lives.

We make use of the IF reward structure as follows. We call \emph{score} the agent's total (undiscounted) reward since the beginning of an episode, remarking that the term extends beyond game-like domains. At time step $t$, the score $u_t$ is
\begin{equation*}
    u_t := \sum_{i=0}^{t-1} r_i .
\end{equation*}
In IF domains, where the score reflects the agent's progress, it is reasonable to treat it as a state variable. We propose maintaining a separate action-value function for each possible score. This action-value function is denoted $Q(h_t, a_t, u_t)$. We call this approach \emph{score contextualisation}.
The use of additional context variables has by now been demonstrated in a number of settings (\citet{DBLP:journals/corr/abs-1903-08254}; \citet{icarte18using}; \citet{DBLP:journals/corr/abs-1711-09874}). First, credit assignment becomes easier since the score provides clues as to the hidden state. Second, in settings with function approximation we expect optimization to be simpler since for each $u$, the function $Q(\cdot, \cdot, u)$ needs only be trained on a subset of the data, and hence can focus on features relevant to this part of the environment. 

In a deep network, we implement score contextualisation using $K$ network heads and a map $\mathcal{J} : \mathbb{N} \to \{ 1, \dots, K \}$ such that the $\mathcal{J}(u_t)^{th}$ head is used when the agent has received a score of $u_t$ at time $t$. This provides the flexibility to either map each score to a separate network head, or multiple scores to one head. Taking $K=1$ uses one monolothic network for all subtasks, and fully relies on this network to identify state from feedback. In our experiments, we assign scores to networks heads using a round-robin scheme with a fixed $K$.
Using \citet{Narasimhan_2015}'s terminology, our architecture consists of a shared \emph{representation generator} $\Phi_R$ with $K$ independent LSTM heads, followed by a feed-forward \emph{action scorer} $\Phi_A(i)$ which outputs the action-values (Figure \ref{fig:memento_architecture}).

\subsection{Action Gating Based on Admissibility}
\label{aux_classifier}

In this section we revisit the idea of using the admissibility function to eliminate or more generally \emph{gate} actions. Consider an action $a$ which is inadmissible in state $s$. By definition, taking this action does not affect the state. We further assume that inadmissible actions produce a constant level of reward, which we take to be 0 without loss of generality:
\begin{equation*}
    a \text{ inadmissible in } s \implies r(s,a) = 0 .
\end{equation*}
This assumption is reasonable in IF domains, and more generally holds true in domains that exhibit subtask structure, such as the video game \textsc{Montezuma's Revenge} \citep{NIPS2016_6383}.
We can combine knowledge of $P$ and $r$ for inadmissible actions with Bellman's equation \eqnref{bellman_equation} to deduce that for any policy $\pi$,
\begin{equation}
    a \text{ inadmissible in } s \implies Q^\pi(s,a) \le \max_{a' \in \cA} Q^\pi(s, a')  \label{eqn:inadmissible_q}
\end{equation}
If we know that $a$ is inadmissible, then we do not need to learn its action-value.

We propose learning a classifier whose purpose is to predict the admissibility function. Given a history $h$, this classifier outputs, for each action $a$, the probability $\hatxi(h, a)$ that this action is admissible. Because of state aliasing, this probability is in general strictly between 0 and 1; furthermore, it may be inaccurate due to approximation error. We therefore consider action gating schemes that are sensitive to intermediate values of  $\hatxi(h, a)$. The first two schemes produce an approximately admissible set $\hat \cA_t$
which varies from time step to time step; the third directly uses the definition of admissibility in a history-based implementation of the consistent Bellman operator.

{\bf Dropout.} The \emph{dropout} method randomly adds each action $a$ to $\hatcAt$ with probability $\hatxi(h_t, a)$. 

{\bf Masking.} The \emph{masking} method uses an elimination threshold $c \in [0, 1)$. The set $\hatcAt$ contains all actions $a$ whose estimated admissibility is at least $c$:
\begin{equation*}
    \hatcAt := \{ a : \hatxi(h_t, a) \ge c \} .
\end{equation*}
The masking method is a simplified version of \citet{Zahavy:2018:LLA:3327144.3327274}'s action elimination algorithm, whose threshold is adaptively determined from a confidence interval, itself derived from assuming a value function and admissibility functions that can be expressed linearly in terms of some feature vector.

In both the dropout and masking methods, we use the action set $\hat \cA_t$ in lieu of the the full action set $\cA$ when selecting exploratory actions.

{\bf Consistent Q-learning for histories (CQLH).} The third method leaves the action set unchanged, but instead drives the action-values of purportedly inadmissible actions to 0. This is done by adapting the consistent Bellman operator \eqnref{consistent_bellman_operator} to the history-based setting. First, we replace the indicator $\indic{s_{t+1} \ne s_t}$ by the probability $\hatxi_t := \hatxi(h_t, a_t)$. Second, we drive $Q(s_t, a_t)$ to $0$ in the case when we believe the state is unchanged, following the argumentation of \eqnref{inadmissible_q}. This yields a version of consistent Q-learning which is adapted to histories, and makes use of the predicted admissibility:

\begin{align}
    \delta^{\textsc{cqlh}}_t = \; &r_t + \gamma \max_{a \in \cA} Q(h_{t+1}, a) \hatxi_t \nonumber \\
    & +\gamma Q(h_t, a_t) (1-\hatxi_t) - Q(h_t, a_t) . \label{eqn:consistent_q_learning_histories}
\end{align}

One may ask whether this method is equivalent to a belief-state average of consistent Q-learning when  $\hatxi(h_t, a_t)$ is accurate, i.e. equals $\xi(h_t, a_t)$. In general, this is not the case: the admissibility of an action depends on the hidden state, which in turns influences the action-value at the next step. As a result, the above method may underestimate action-values when there is state aliasing (e.g., $\hatxi(h_t, a_t) \approx 0.5$), and yields smaller action gaps than the state-based version when $\hatxi(h_t, a_t) = 1$. However, when $a_t$ is known to be inadmissible ($\hatxi(h_t, a_t) = 0$), the methods do coincide, justifying its use as an action gating scheme.

We implement these ideas using an \textit{auxiliary classifier} $\Phi_C$. For each action $a$, this classifier outputs the estimated probability $\hatxi(h_t,a)$, parametrized as a sigmoid function. These probabilities are learned from bandit feedback: after choosing $a$ from history $h_t$, the agent receives a binary signal $e_t$ as to whether $a$ was admissible or not. In our setting, learning this classifier is particularly challenging because the agent must predict admissibility solely based on the history $h_t$. As a point of comparison, using the information-gathering commands \textsc{look} and \textsc{inventory} to establish the state, as proposed by \citet{Zahavy:2018:LLA:3327144.3327274}, leads to a simpler learning problem, but one which does not consider the full history. The need to learn $\hatxi(h_t, a)$ from bandit feedback also encourages methods that generalize across histories and textual descriptions.

\section{A Synthetic IF Benchmark}

Both score contextualisation and action gating are tailored to domains that exhibit the structure typical of interactive fiction. To assess how useful these methods are, we will make use of a synthetic benchmark based on the TextWorld framework \citep{cote18textworld}. TextWorld provides a reinforcement learning interface to text-based games along with an environment specification language for designing new environments. Environments provide a set of locations, or \emph{rooms}, objects that can picked up and carried between locations, and a reward function based on interacting with these objects. Following the genre, special \emph{key} objects are used to access parts of the environment.

Our benchmark provides seven environments of increasing complexity, which we call \emph{levels}. We control complexity by adding new rooms and/or objects to each successive level. Each level also requires the agent to complete a number of subtasks (Table \ref{sample-table}), most of which involve carrying one or more items to a particular location. Reward is provided only when the agent completes one of these subtasks. Thematically, each level involves collecting food items to make a salad, inspired by the first TextWorld competition. Example objects include an apple and a head of lettuce, while example actions include \texttt{get apple} and \texttt{slice lettuce with knife}. Accordingly we call our benchmark SaladWorld.

SaladWorld provides a graded measure of an agent architecture's ability to deal with both partial observability and large action spaces. Indeed, completing each subtasks requires memory of what has previously been accomplished, along with where different objects are. Together with this, each level in the SaladWorld involves some amount of history-dependent admissibility i.e the admissibility of the action depends on the history rather than the state. For example, \texttt{put lettuce on counter} can only be accomplished once \texttt{take lettuce} (in a different room) has happened. Keys pose an additional difficulty as they do not themselves provide reward. As shown in Table \ref{sample-table}, the number of possible actions rapidly increases with the number of objects in a given level. Even the small number of rooms and objects considered here preclude the use of tabular representations, as the state space for a given level is the exponentially-sized cross-product of possible object and agent locations. In fact, we have purposefully designed SaladWorld as a small challenge for IF agents, and even our best method falls short of solving the harder levels within the allotted training time. Full details are given in Table \ref{info-table} in the appendix.

\begin{table}[t]
\caption{Main characteristics of each level in our synthetic benchmark.}
\label{sample-table}

\begin{center}
\begin{small}
\begin{sc}
\begin{tabular}{lccccr}
\toprule
Level & \# Rooms & \# Objects & \# Sub-tasks & $|\mathcal{A}|$ \\
\midrule
1 & 4 & 2 & 2 & 8\\
2 & 7 & 4 & 3 & 15\\
3 & 7 & 4 & 3 & 15\\
4 & 9 & 8 & 4 & 50\\
5 & 11 & 15 & 5 & 141\\
6 & 12 & 20 & 6 & 283\\
7 & 12 & 20 & 7 & 295\\
\bottomrule
\end{tabular}
\end{sc}
\end{small}
\end{center}

\end{table}

\begin{table*}[!ht]
\caption{Subtasks information and scores possible for each level of the suite.}
\label{info-table}

\centering
\begin{tabular}{|p{1cm}|p{12cm}|p{3cm}|}
\toprule
Level & Subtasks & Possible Scores \\
\midrule
1   & Following subtasks with reward and fulfilling condition:\begin{itemize}
    \item 10 points when the agent first enters the vegetable market. 
    \item 5 points when the agent gets lettuce from the vegetable market and puts lettuce on the counter. 
\end{itemize}& 10, 15          \\
2   &  All subtasks from previous level plus this subtask: \begin{itemize}
    \item 5 points when the agent takes the blue key from open space, opens the blue door, gets tomato from the supermarket and puts it on the counter in the kitchen.
\end{itemize}& 5, 10, 15, 20                \\
3   &  All subtasks from level 1 plus this subtask: \begin{itemize}
    \item 5 points when the agent takes the blue key from open space, goes to the garden, opens the blue door with the blue key, gets tomato from the supermarket and puts it on the counter in the kitchen.
\end{itemize} 
\textbf{Remark:} Level 3 game differs from Level 2 game in terms of number of steps required to complete the additional sub-task (which is greater in case of Level 3) 
& 5, 10, 15, 20                \\
4   &  All subtasks from previous level plus this subtask: \begin{itemize}
    \item 5 points when the agent takes parsley from the backyard and knife from the cutlery shop to the kitchen, puts parsley into fridge and knife on the counter. 

\end{itemize} & 5, 10, 15, 20, 25                \\
5   &  All subtasks from previous level plus this subtask: \begin{itemize}
    \item 5 points when the agent goes to fruit shop, takes chest key, opens container with chest key, takes the banana from the chest and puts it into the fridge in the kitchen. 

\end{itemize} & 5, 10, 15, 20, 25, 30               \\
6   &  All subtasks from previous level plus this subtask: \begin{itemize}
    \item 5 points when the agent takes the red key from the supermarket, goes to the playroom, opens the red door with the red key, gets the apple from cookhouse and puts it into the fridge in the kitchen.

\end{itemize} & 5, 10, 15, 20, 25, 30, 35             \\
7   &  All subtasks from previous level plus this subtask: \begin{itemize}
    \item 5 points when the agent prepares the meal.

\end{itemize} & 5, 10, 15, 20, 25, 30, 35, 40           \\
\bottomrule
\end{tabular}
\end{table*}

\section{Empirical Analysis}

In the first set of experiments, we use SaladWorld to establish that both score contextualisation and action gating provide positive benefits in the context of IF domain. We then validate these findings on the celebrated text-based game \textsc{Zork} used in prior work \cite{Fulda_2017,Zahavy:2018:LLA:3327144.3327274}.

Our baseline agent is the LSTM-DRQN agent \citep{yuan2018counting} but with a different action representation. We augment this baseline with either or both score contextualisation and action gating, and observe the resulting effect on agent performance in SaladWorld. We measure this performance as the fraction of subtasks completed during an episode, averaged over time. In all cases, our results are generated from 5 independent trials of each condition. To smooth the results, we use moving average with a window of 20,000 training steps. The graphs and the histograms report average $\pm$ std. deviation across the trials. 

Score contextualisation uses $K=5$ network heads; the baseline corresponds to $K=1$. Each head is trained using the Adam optimizer \citep{kingma:adam} with a learning rate $\alpha=0.001$ to minimize a Q-learning loss \citep{mnih2015human} with a discount factor of $\gamma = 0.9$. The auxiliary classifier $\Phi_C$ is trained with the binary cross-entropy loss over the selected action's admissibility (recall that our agent only observes the admissibility function for the selected action). Training is done using a balanced form of prioritized replay which we found improves baseline performance appreciably. Specifically, we use the sampling mechanism described in \citet{SDMIA15-Hausknecht} with prioritization i.e we sample $\tau_p$ fraction of episodes that had atleast one positive reward, $\tau_n$ fraction with atleast one negative reward and $1 - \tau_p - \tau_n$ from whole episodic memory $\mathcal{D}$. Section \ref{baseline_ablation} in the Appendix compares the baseline agent with and without prioritization. For prioritization, $\tau_p = \tau_n = 0.25$.

Actions are chosen from the estimated admissible set $\hatcAt$ according to an $\epsilon$-greedy rule, with $\epsilon$ annealed linearly from 1.0 to 0.1 over the first million training steps. To simplify exploration, our agent further takes a forced \textsc{look} action every 20 steps. Each episode lasts for a maximum $T$ steps. For Level $1$ game, $T=100$, whereas for rest of the levels $T=200$.To simplify exploration, our agent further takes a forced \textsc{look} action every 20 steps (Section \ref{baseline_ablation}).Full details are given in the Appendix (Section \ref{sec:training_details}).

\begin{figure}[htb]
    \centering
    \includegraphics[scale=0.1]{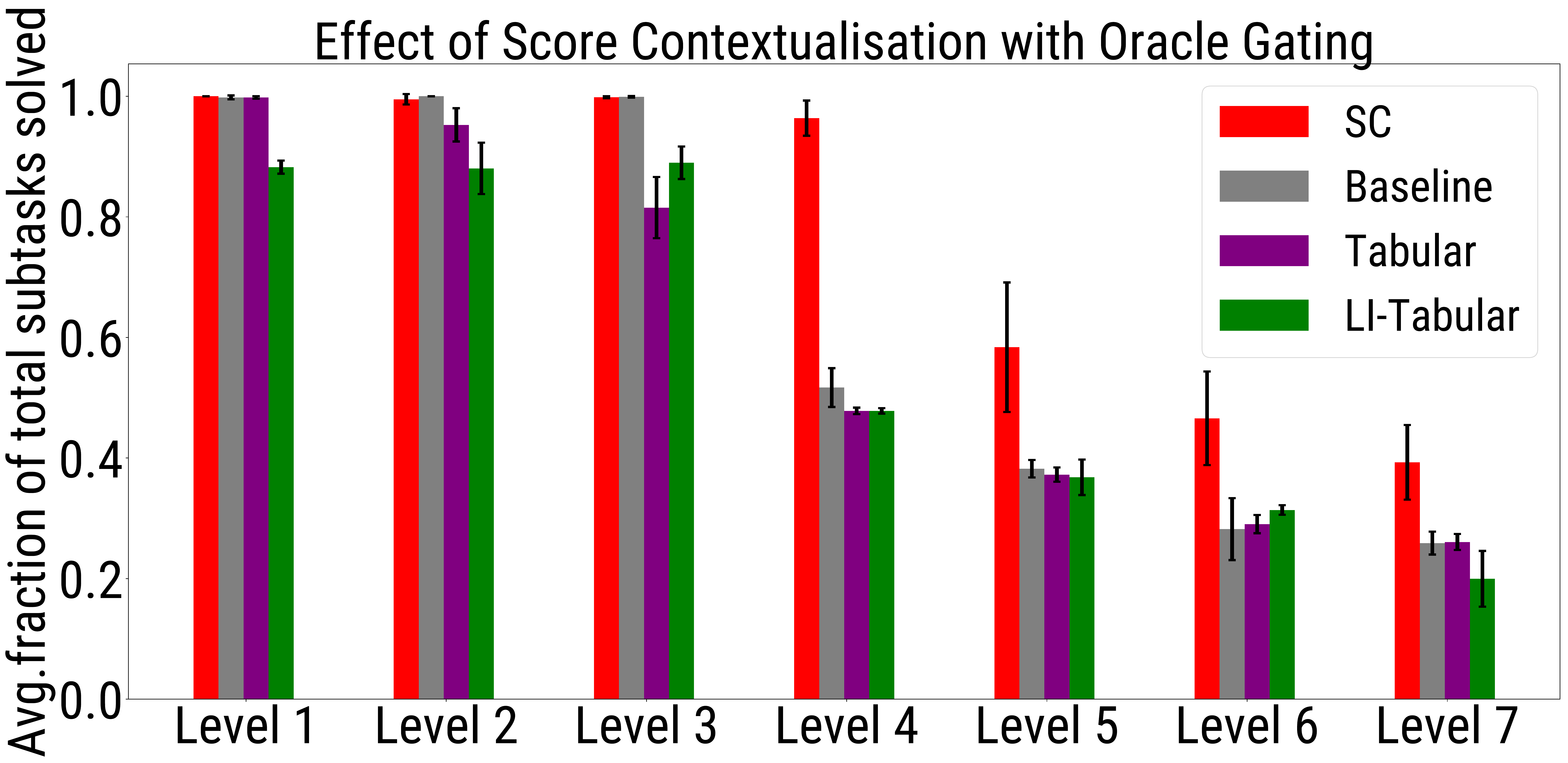}
    \caption{Fraction of tasks solved by each method at the end of training for 1.3 million steps. The tabular agents, which do not take history into account, perform quite poorly. LI stands for ``look, inventory'' (see text for details).}
    \label{fig:adm_images}
\end{figure}

\subsection{Score Contextualisation}

\begin{figure}[!htb]
    \centering %
  \includegraphics[width=\linewidth]{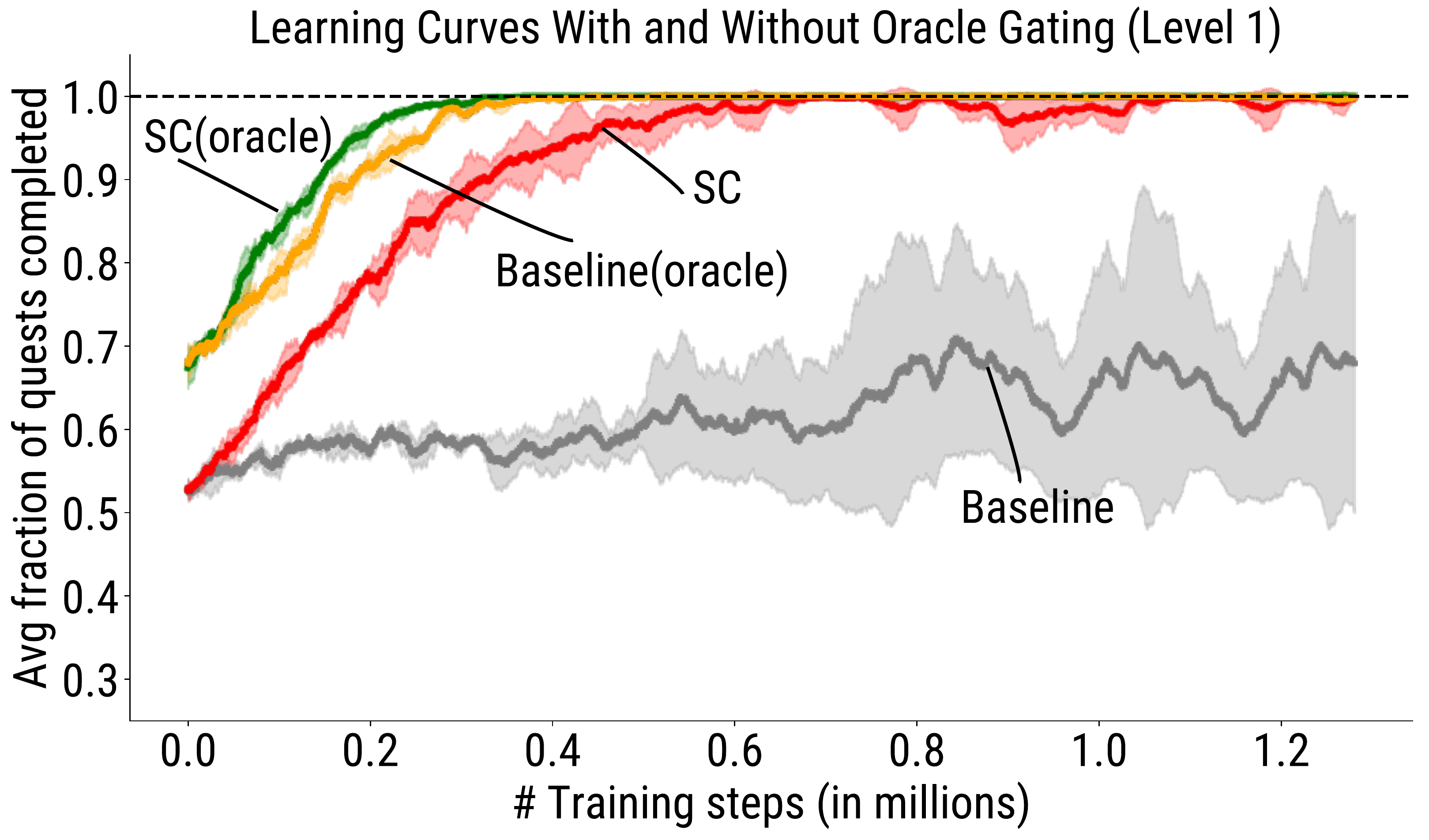}
  \includegraphics[width=\linewidth]{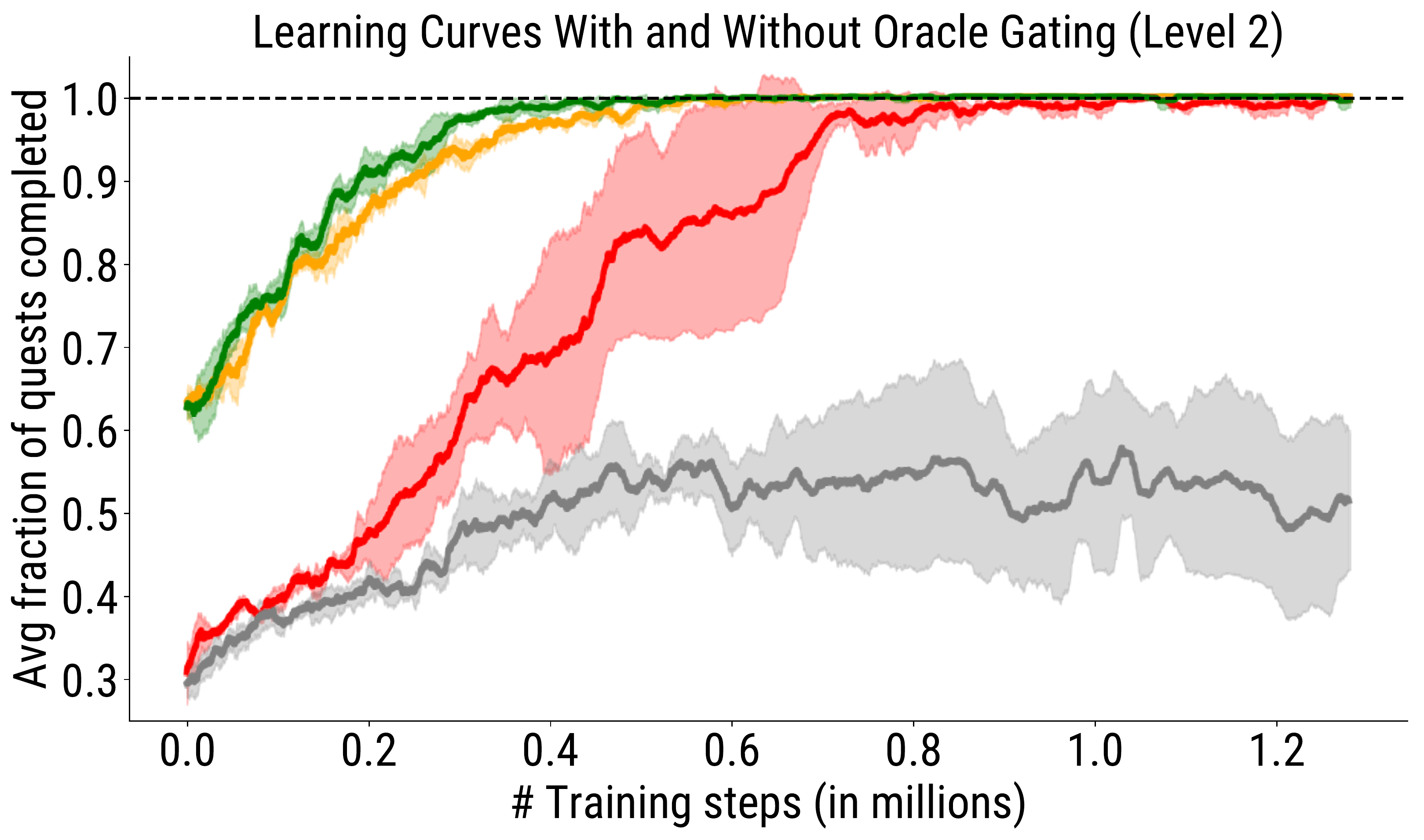}
\caption{Comparing whether score contextualisation as an architecture provides a useful representation for learning to act optimally. Row 1 and 2 correspond to Level 1 and 2 respectively.}
\label{fig:simple_images}
\end{figure}

We first consider the effect of score contextualisation on our agents' ability to complete tasks in SaladWorld. We ask,
\begin{quote}
    \textit{Does score contextualisation mitigate the negative effects of partial observability?}
\end{quote}
We begin in a simplified setting where the agent knows the admissible set $\cA_t$. We call this setting \emph{oracle gating}. This setting lets us focus on the impact of contextualisation alone. We compare our score contextualisation (SC) to the baseline and also to two ``tabular'' agents. The first tabular agent treats the most recent feedback as state, and hashes each unique description-action pair to a Q-value. This results in a memoryless scheme that ignores partial observability.
The second tabular agent performs the information-gathering actions \textsc{look} and \textsc{inventory} to construct its state description, and also hashes these to unique Q-values. Accordingly, we call this the ``LI-tabular'' agent. This latter scheme has proved to be a successful heuristic in the design of IF agents \citep{Fulda_2017}, but can be problematic in domains where taking information-gathering actions can have negative consequences (as is the case in \textsc{Zork}).

Figure \ref{fig:adm_images} shows the performance of the four methods across SaladWorld levels, after 1.3 million training steps. We observe that the tabular agents' performance suffers as soon as there are multiple subtasks, as expected. The baseline agent performs well up to the third level, but then shows significantly reduced performance. We hypothesize that this occurs because the baseline agent must estimate the hidden state from longer history sequences and effectively learn an implicit contextualisation. Beyond the fourth level, the performance of all agents suffers, suggesting the need for a better exploration strategy, for example using expert data \cite{DBLP:journals/corr/abs-1905-09700}.

We find that score contextualisation performs better than the baseline when the admissible set is unknown. Figure \ref{fig:simple_images} compares learning curves of the SC and baseline agents with oracle gating and using the full action set, respectively, in the simplest of levels (Level 1 and 2). We find that score contextualisation can learn to solve these levels even without access to $\cA_t$, whereas the baseline cannot. Our results also show that oracle gating simplifies the problem, and illustrate the value in handling inadmissible actions differently.%

We hypothesize that score contextualisation results in a simpler learning problem in which the agent can more easily learn to distinguish which actions are relevant to the task, and hence facilitate credit assignment. Our result indicates that it might be unreasonable to expect contextualisation to arise naturally (or easily) in partially observable domains with large actions sets. We conclude that score contextualisation mitigates the negative effects of partial observability.

\begin{figure}[!htb]
    \centering
    \includegraphics[scale=0.1]{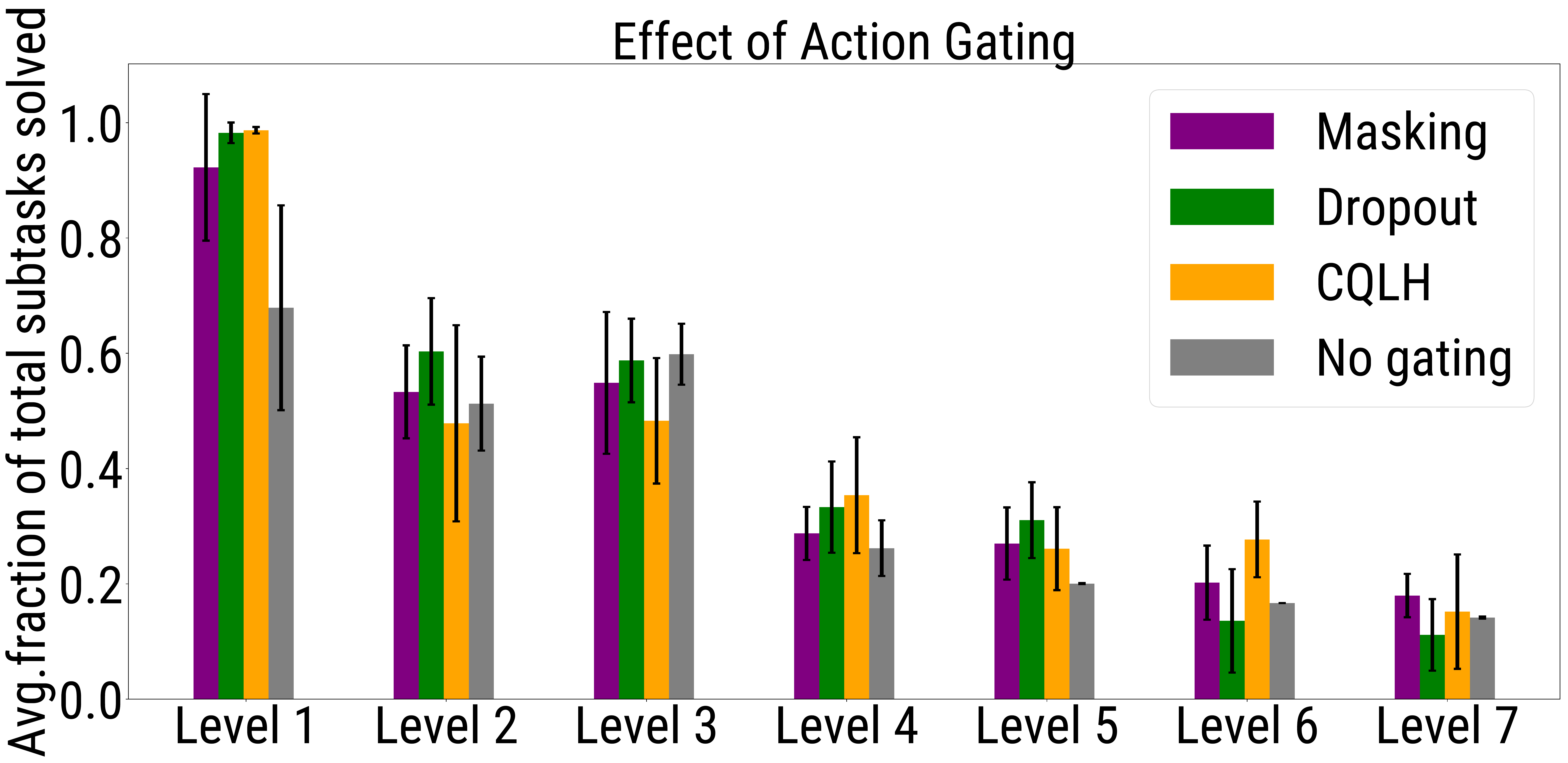}
    \caption{Fraction of tasks solved by each method at the end of training for 1.3 million steps. Except in Level 1, action gating by itself does not improve end performance.}
    \label{fig:action_gating_histogram}
\end{figure}

\begin{figure}[htb]
    \centering %
  \includegraphics[width=\linewidth]{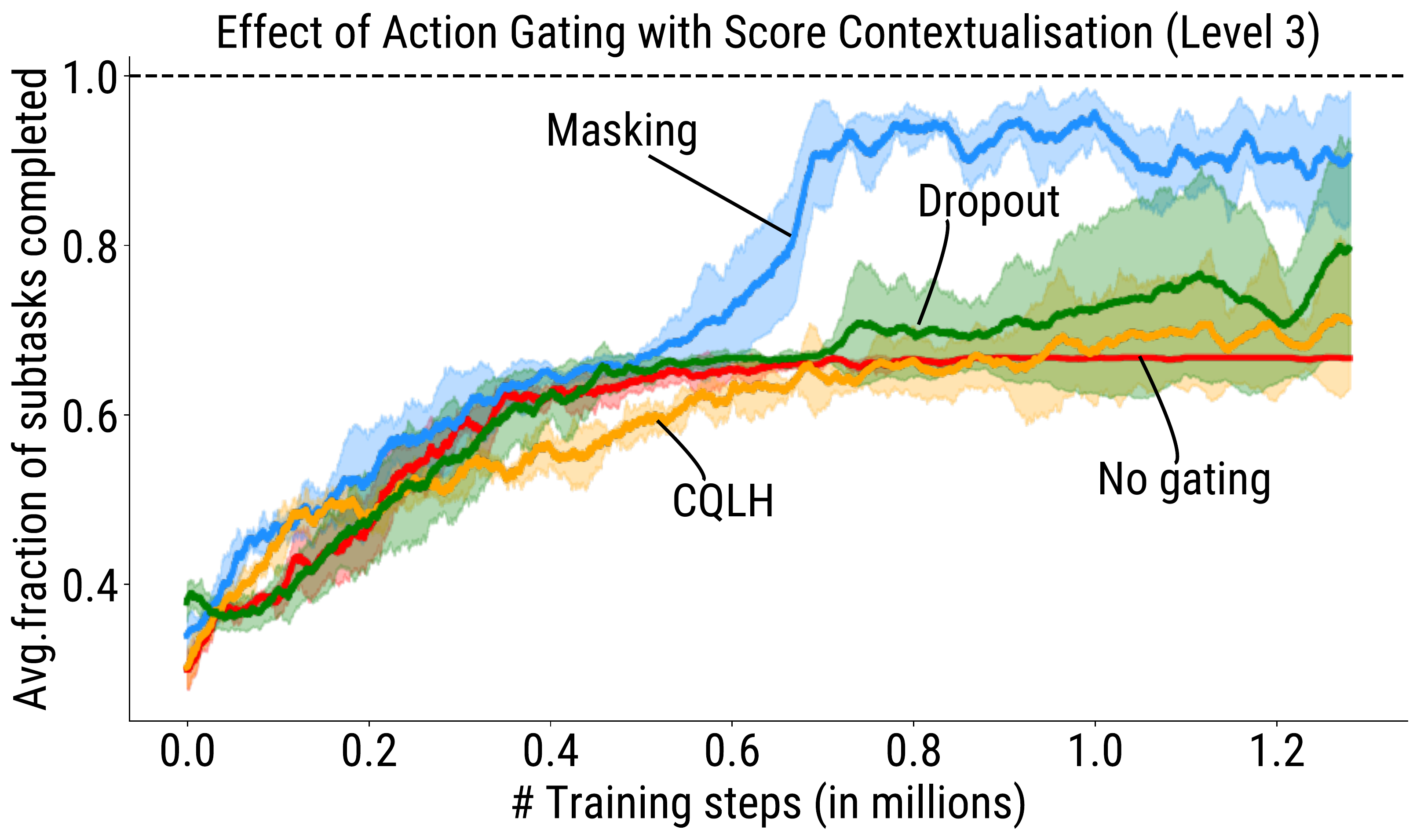}
\caption{Effectiveness of action gating with score contextualisation in Level 3. Of the three methods, masking performs best.}
\label{fig:level_3_images}
\end{figure}

\subsection{Score Contextualisation with Learned Action Gating}

The previous experiment (in particular, Figure \ref{fig:simple_images}) shows the value of restricting action selection to admissible actions.
With the goal in mind of designing an agent that can operate from feedback alone, we now ask:
\begin{quote}
    \textit{Can an agent learn more efficiently when given bandit feedback about the admissibility of its chosen actions?}
\end{quote}
We address this question by comparing our three action gating mechanisms. As discussed in Section \ref{aux_classifier}, the output of the auxiliary classifier describes our estimate of an action's admissibility for a given history.

As an initial point of comparison, we tested the performance of the baseline agent when using the auxiliary classifier's output to gate actions. For the masking method, we selected $c=0.001$ from a larger initial parameter sweep. The results are summarized in Figure \ref{fig:action_gating_histogram}. While action gating alone provides some benefits in the first level, performance is equivalent for the rest of the levels.

\begin{figure}
    \centering
    \includegraphics[scale=0.1]{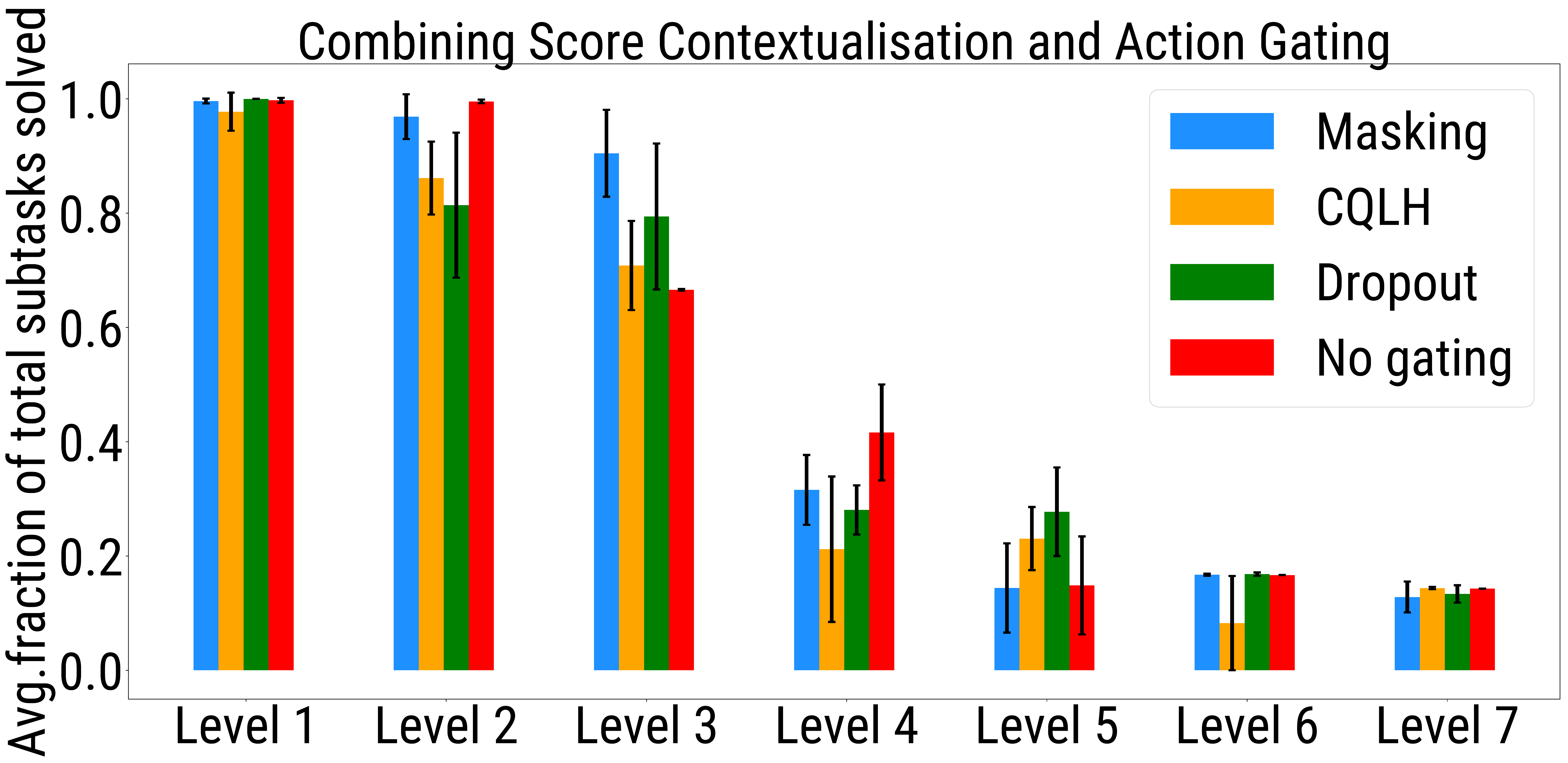}
    \caption{Fraction of tasks solved by each method at the end of training for 1.3 million steps. For first 3 levels, SC + Masking is better or equivalent to SC. For levels 4 and beyond, better exploration strategies are required. }
\label{fig:memento_action_gating_histogram}
\end{figure}

However, when combined with score contextualisation (see Fig \ref{fig:level_3_images}, \ref{fig:memento_action_gating_histogram}), we observe some performance gains. In Level 3 in particular, we almost recover the performance of the SC agent with oracle gating. From our results we conclude that masking with the right threshold works best, but leave as an open question whether the other action gating schemes can be improved.

\begin{figure}
    \centering
    \includegraphics[scale=0.1]{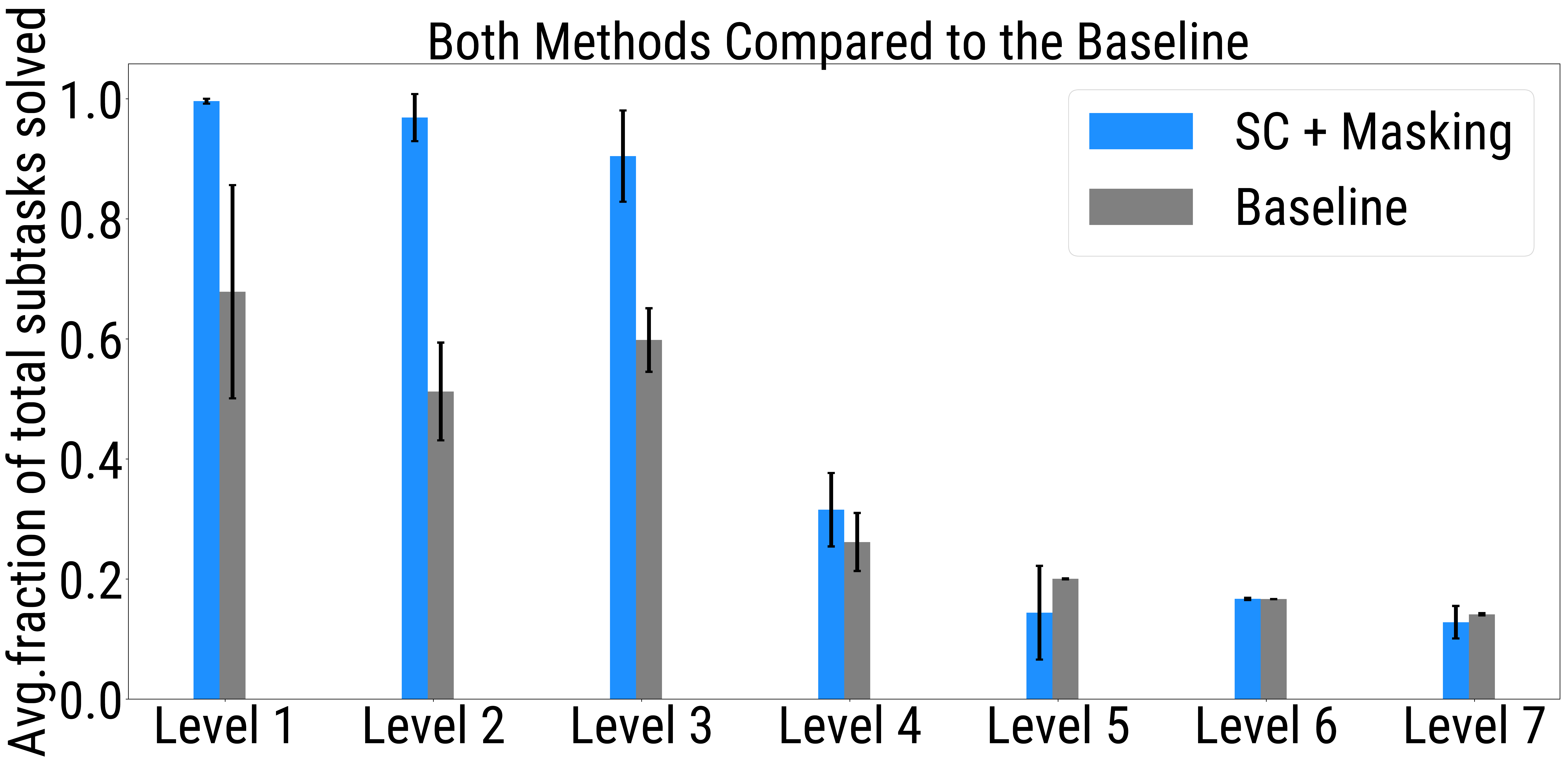}
    \caption{Score contextualisation and masking compared to the baseline agent. We show the fraction of tasks solved by each method at the end of training for 1.3 million steps.}
    \label{fig:memento_mask_baseline_gating_histogram}
\end{figure}

Figure \ref{fig:memento_mask_baseline_gating_histogram} shows the final comparison between the baseline LSTM-DRQN and our new agent architecture which incorporates action gating and score contextualisation (full learning curves are provided in the appendix, Figure \ref{fig:all_learning_curves_final_comparison}). Our results show that the augmented method significantly outperforms the baseline, and is able to handle more complex IF domains. From level 4 onwards, the learning curves in the appendix show that combining score contextualisation with masking results in faster learning, even though final performance is unchanged. We posit that better exploration schemes are required for further progress in SaladWorld.

\subsection{Zork}

As a final experiment, we evaluate our agent architecture on the interactive fiction \textsc{Zork} I, the first installment of the popular trilogy. \textsc{Zork} provides an interesting point of comparison for our methods, as it is designed by and for humans -- following the ontology of \citet{Bellemare_2013}, it is a domain which is both \emph{interesting} and \emph{independent}. Our main objective is to compare the different methods studied with \citet{Zahavy:2018:LLA:3327144.3327274}'s AE-DQN agent. Following their experimental setup, we take $\gamma = 0.8$ and train for 2 million steps. All agents use the smaller action set (131 actions). Unlike AE-DQN, however, our agent does not use information-gathering actions (\textsc{look} and \textsc{inventory}) to establish the state.

\begin{figure}
    \centering
    \includegraphics[scale=0.25]{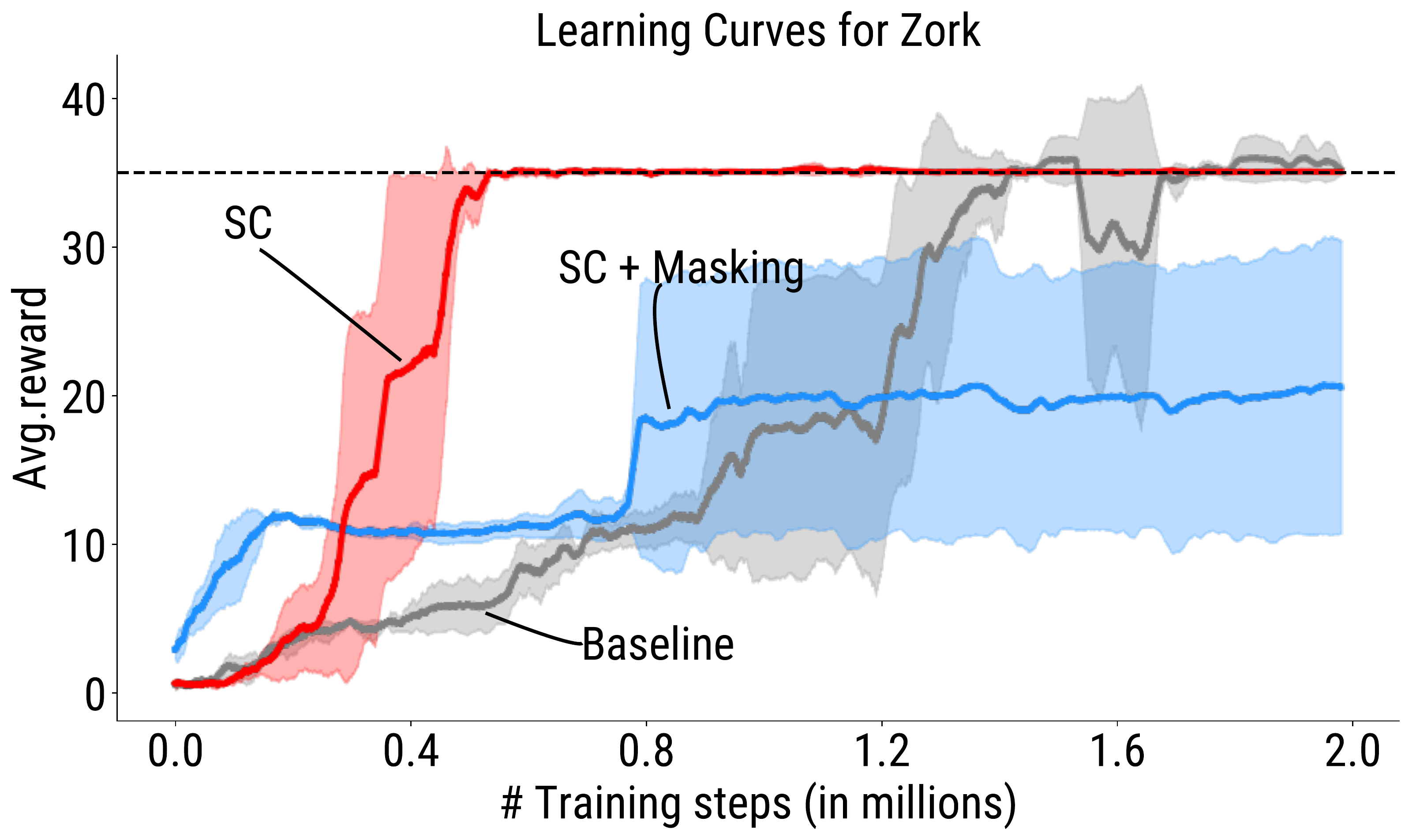}
    \caption{Learning curves for different agents in Zork.}
    \label{fig:zork_compare}
\end{figure}

Figure \ref{fig:zork_compare} shows the corresponding learning curves. Despite operating in a harder regime than AE-DQN, the score contextualizing agent reaches a score comparable to AE-DQN, in about half of the training steps. All agents eventually fail to pass the 35-point benchmark, which corresponds to a particularly difficult in-game task (the ``troll quest'') which involves a timing element, and we hypothesize requires a more intelligent exploration strategy.

  \section{Related Work}
\label{related_work}
{\bfseries RL applied to Text Adventure games:}
LSTM-DQN by \citet{Narasimhan_2015} deals with parser-based text adventure games and uses an LSTM to generate feedback representation. The representation is then used by an action scorer to generate scores for the action verb and objects. The two scores are then averaged to determine Q-value for the state-action pair. In the realm of choice-based games, \citet{he-etal-2016-deep}  uses two separate deep neural nets to generate representation for feedback and action respectively. Q-values are calculated by dot-product of these representations. None of the above approaches deals with partial observability in text adventure games. 

{\bfseries Admissible action set learning:}
\citet{tao2018solving} approach the issue of learning admissible set given context as a supervised learning one. They train their model on (input, label) pairs where input is context (concatenation of feedbacks by \textsc{look} and \textsc{inventory}) and label is the list of admissible commands given this input. AE-DQN  \citep{Zahavy:2018:LLA:3327144.3327274} employs an additional neural network to prune in-admissible actions from action set given a state. Although the paper doesn't deal with partial observability in text adventure games, authors show that having a tractable admissible action set led to faster convergence. \citet{Fulda_2017} work on bounding  the  action set through  affordances. Their agent is trained through tabular Q-Learning.

{\bfseries Partial Observability:}
 \citet{yuan2018counting} replace the shared MLP in \citet{Narasimhan_2015} with an LSTM cell to calculate context representation. However, they use concatenation of feedbacks by \textsc{look} and \textsc{inventory} as the given state to make the game more observable. Their work also doesn't focus on pruning in-admissible actions given a context. Finally, \citet{ammanabrolu-riedl-2019-playing} deal with partial observability by representing state as a knowledge graph and continuously updating it after every game step. However, the graph update rules are hand-coded; it would be interesting to see they can be learned during gameplay.

\section{Conclusions and Future work}

We introduced two algorithmic improvements for deep reinforcement learning applied to interactive fiction (IF). While naturally rooted in IF, we believe our ideas extend more generally to partially observable domains and large discrete action spaces. Our results on  SaladWorld and \textsc{Zork} show the usefulness of these improvements. Going forward, we believe better contextualisation mechanisms should yield further gains. In \textsc{Zork}, in particular, we hypothesize that going beyond the 35-point limit will require more tightly coupling exploration with representation learning.

\section{Acknowledgments}
This work was funded by the CIFAR Learning in Machines and Brains program. Authors thank Compute Canada for providing the computational resources.
\bibliographystyle{aaai}
\bibliography{main}

\clearpage
\appendix

\section{Training Details}\label{sec:training_details}
\subsection{Hyper-parameters}
{\bfseries Training hyper-parameters:} For all the experiments unless specified, $\gamma=0.9$. Weights for the learning agents are updated every $4$ steps. Agents with score contextualisation architecture have $K=5$ network heads. Parameters of score contextualisation architecture are learned end to end with Adam optimiser \citep{kingma:adam} with learning rate $\alpha=0.001$. To prevent imprecise updates for the initial states in the transition sequence due to in-sufficient history, we use updating mechanism proposed by \citet{lample2017playing}. In this mechanism, considering the transition sequence of length $l, o_1, o_2, \dots, o_l$, errors from $o_1, o_2, \dots, o_n$ aren't back-propagated through the network. In our case, the sequence length $l=15$ and minimum history size for a state to be updated $n=6$ for all experiments.  Score contextualisation heads are trained to minimise the Q-learning loss over the whole transition sequence. On the other hand, $\Phi_C$ minimises the BCE (binary cross-entropy) loss over the predicted admissibility probability and the actual admissibility signal for every transition in the transition sequence. The behavior policy during training is $\epsilon-$greedy over the admissible set $\hatcAt$. Each episode lasts for a maximum $T$ steps. For Level $1$ game, we anneal $\epsilon = 1$ to $0.1$ over $1000000$ steps and $T=100$. For rest of the games in the suite, we anneal $\epsilon = 1$ to $0.1$ over $1000000$ steps and $T=200$.

{\bfseries Architectural hyper-parameters:} In $\Phi_R$, word embedding size is $20$ and the number of hidden units in encoder LSTM is $64$. For a network head $k$, the number of hidden units in context LSTM is $512$; $\Phi_A(k)$ is a two layer MLP: sizes of first  and second layer are 128 and $|\mathcal{A}|$ respectively. $\Phi_C$ has the same configuration as $\Phi_A(k)$.

\begin{figure}
    \centering
    \includegraphics[width=\linewidth]{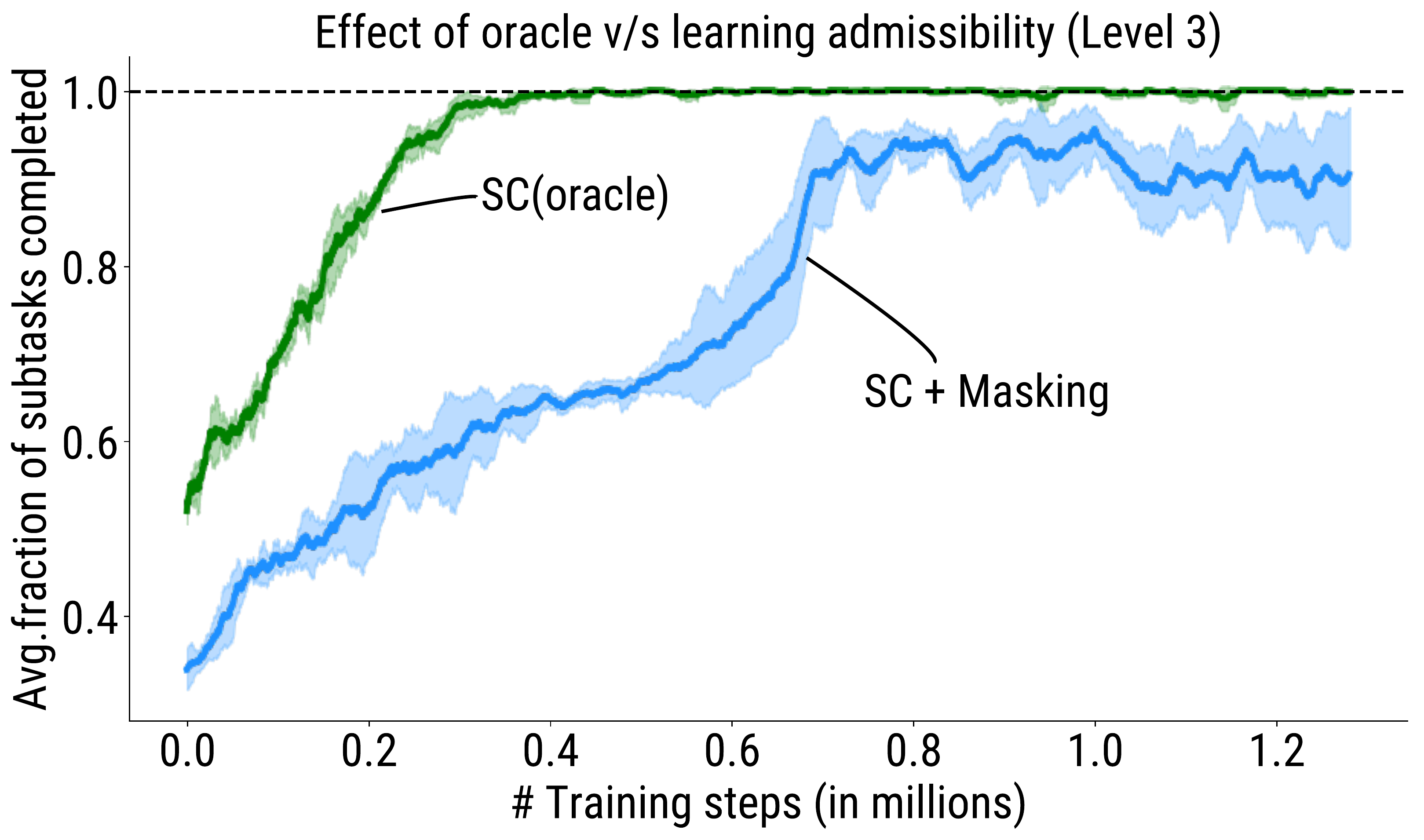}
    \caption{Comparing the effect of oracle gating versus learning admissibility from bandit feedback. Learning is faster in case of oracle gating since agent is given admissible action set resulting in overall better credit assignment.}
    \label{fig:memento_mask_oracle}
\end{figure}

\begin{figure}
    \centering
    \includegraphics[width=\linewidth]{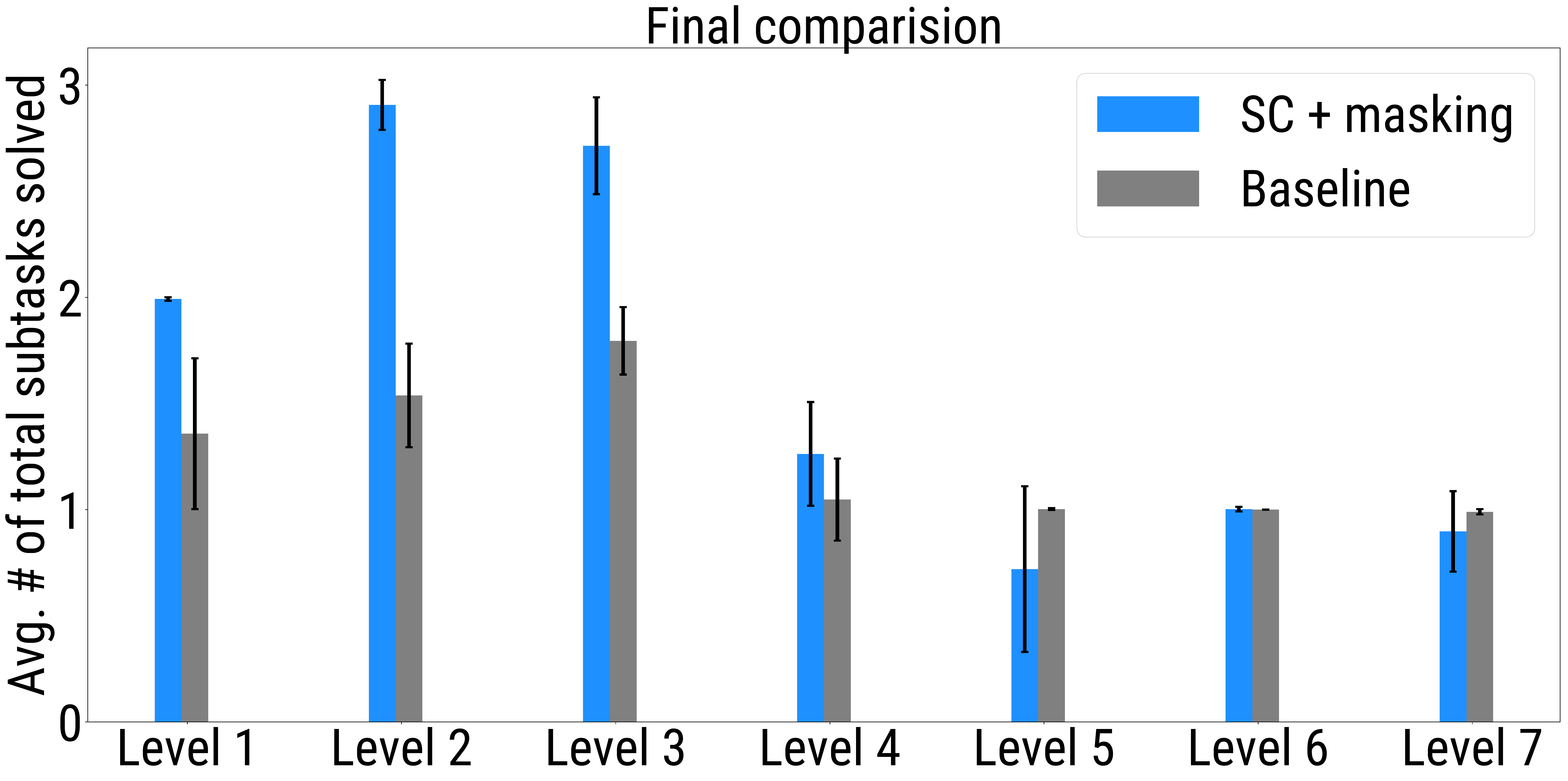}
    \caption{Final comparison shows that our algorithmic enhancements improve the baseline. We show number of tasks solved by each method at the end of training for 1.3 million steps.}
    \label{fig:memento_mask_baseline_gating_histogram_alternate}
\end{figure}

 \subsection{Action Gating Implementation}\label{action_gating_implementation}
For \emph{dropout} and \emph{masking} when selecting actions, we set $Q(h_t, a_t) = -\infty$ for $a \notin \hatcAt$. Since $\hatxi_t$ is basically an estimate for admissibility for action $a$ given history $h_t$, we use (\ref{eqn:consistent_q_learning_histories}) to implement consistent Q value backups: $Q(h_t, a_t) \gets Q(h_t, a_t) + \alpha \delta''_t$ where

\begin{multline*}
  \delta''_t = r_t + \gamma \big[\max_{a \in \cA} Q(h_{t+1}, a) \hatxi_t + \\ Q(h_{t+1}, a_t) (1-\hatxi_t)\big]- Q(h_t, a_t)  
\end{multline*}

We notice by using the above equation, that for an action $a$ inadmissible in $s$, it's value indeed reduces to $0$ over time. 

\subsection{Baseline Modifications}
We modify LSTM-DRQN \citep{yuan2018counting} in two ways. First, we concatenate the representations  $\Phi_R(o_t)$ and $\Phi_R(a_{t-1})$ before sending it to the history LSTM, in contrast \citet{yuan2018counting} concatenates the inputs $o_t$ and $a_{t-1}$ first and then generates $\Phi_R([o_t;a_{t-1}])$. Second, we modify the action scorer as action scorer in the LSTM-DRQN could only handle commands with two words.  
\section{Notations and Algorithm}
Following are the notations important to understand the algorithm:
\begin{itemize}
    \item $o_t, r_t, e_t:$ observation (i.e. feedback), reward and admissibility signal received at time $t$.
    \item $a_{t}:$ command executed in game-play at time $t$.
    \item $u_{t}:$ cumulative rewared/score at time $t$.
    \item $\Phi_R:$ representation generator.
    \item $\Phi_C:$ auxiliary classifier.
    \item $K$ : number of network heads in score contextualisation architecture.
    \item $\mathcal{J}$: dictionary mapping cumulative rewards to network heads.
    \item $H(k):$ LSTM corresponding to network head $k$.
    \item $\Phi_A(k):$ Action scorer corresponding to network head $k$.
    \item $h_t:$ agent's context/history state at time $t$.
    \item $T$: maximum steps for an episode.
    \item $p_i:$ boolean that determines whether +ve reward was received in episode $i$.
    \item $q_i:$ boolean that determines whether -ve reward was received in episode $i$.
    \item $\tau_p:$ fraction of episodes where $\exists t < T : r_t > 0$ 
    \item $\tau_n:$ fraction of episodes where $\exists t < T : r_t < 0$ 
    \item $l:$ sequence length.
    \item $n:$ minimum history size for a state to be updated.
    \item $\mathcal{A}:$ action set.
    \item $\hatcAt:$ admissible set generated at time $t$.
    \item $I_{target}:$ update interval for target network
    \item $\epsilon:$ parameter for $\epsilon-$greedy exploration strategy.
    \item $\epsilon_1:$ softness parameter i.e. $\epsilon_1$ fraction of times $\hatcAt = \mathcal{A}$.
    \item $c:$ threshold parameter for action elimination strategy Masking. 
    \item $G_{\textrm{max}}:$ maximum steps till which training is performed. 
    
\end{itemize}
Full training procedure is listed in Algorithm \ref{memento_algorithm}.

\begin{algorithm*}[!ht]
\caption{General training procedure}\label{memento_algorithm}
\begin{algorithmic}[1]
\Function{ACT$(o_t, a_{t-1}, u_t,h_{t-1},\mathcal{J}, \epsilon, \epsilon_1, c,\theta)$}{}
\State Get network head $k = \mathcal{J}(u_t)$.
\State $h_t \gets$ LSTM $H(k)[w_t,h_{t-1}]$. 
\State ${Q}(h_t, :, u_t; \theta) \gets \Phi_{A}(k)(h_
t); \hatxi(h_t, a; \theta) \gets \Phi_C(h_t)$.
\State Generate $\hatcAt$ (see Section \ref{aux_classifier}).
\State With probability $\epsilon, a_t \gets \textrm{Uniform}(\hatcAt)$, else $a_t \gets \textrm{argmax}_{a\in \hatcAt} {Q}(h_t, a, u_t; \theta)$
\State return $a_t, h_t$
\EndFunction

\Function{TARGETS$(f, \gamma, \theta^{-})$}{}
\State $(a_0,o_1,a_1,r_2, u_2,e_2,o_2,\dots,o_{l},a_{l},r_{l+1},e_{l+1}, u_{l+1})\gets f;h_{b,0}\gets0$
\State Pass transition sequence through $H$ to get $h_{b,1}, h_{b,2}, \dots, h_{b,l}$
\State $E_{b,i} \gets \xi(h_{b,i},a_{i}; \theta^{-})$
\State $y_{b,i} \gets \textrm{max}_{a\in \mathcal{A}} {Q}(h_{b,i+1},a, u_{b,i+1}; \theta^{-})$ 
\State $y_{b,i} \gets E_{b,i} y_{b,i} + (1- E_{b,i})\ {Q}(h_{b,i+1},a_i, u_{b,i+1}; \theta^{-})$ \textbf{if} using CQLH.
\State $y_{b,i} \gets r_{i+1}$ \textbf{if} $o_i$ is terminal \textbf{else} $y_{b,i} \gets r_{i+1} + \gamma y_{b,i}$
\State return $y_{b, :}, E_{b, :}$
\EndFunction
\\\hrulefill
\State {\bfseries Input:} $G_{\textrm{max}}, I_{\text{look}}, I_{\text{update}}, \gamma, \epsilon_1,\epsilon,c, K,\indic{_{\text{using} \Phi_C}}, n$
\State Initialize episodic replay memory $\mathcal{D}$, global step counter $G \gets 0$, dictionary $\mathcal{J}=\{\}$.
\State Initialize parameters $\theta$ of the network, target network parameter $\theta^{-} \gets \theta$.
\While{$G < G_{\textrm{max}}$}  
\parState {Initialize score $u_1=0$, hidden State of $H$, $h_0=0$ and get start textual description $o_1$ and initial command $a_0=\textrm{'look'}$. Set $p_k \gets 0 , q_k \gets 0$.}
\For{$t \gets 1$ to $T$}  
\State  $a_t, h_{t} \gets $ ACT$(o_t, a_{t-1}, u_t,h_{t-1},\mathcal{J}, \epsilon, \epsilon_1, c,\theta)$
\State $a_t \gets$ 'look' \textbf{if} {$t \bmod 20 == 0$}
\State Execute action $a_t$, observe $\{r_{t+1}, o_{t+1}, e_{t+1}\}$.
\State $p_k\gets1$ \textbf{if} $r_t>0$; $q_k\gets1$ \textbf{if} $r_t<0$; $u_{t+1}\gets u_t + r_t$
\State Sample minibatch of transition sequences $f$
\State $y_{b, :}, E_{b, :} \gets $ TARGETS$(f, \gamma, \theta^{-})$
\State Perform gradient descent on $\mathcal{L}(\theta) = \sum_{i=j+n-1}^{j+l}[y_{b,i} - {Q}(h_{b,i}, a_i, u_{b,i}; \theta)^2 + \indic{_{using \Phi_C}} \ \textrm{BCE}(e_i, E_{b, i})]$
\State $\theta^{-} \gets \theta$ \textbf{if} {$t \bmod I_{update} == 0$}
\State $G \gets G + 1$
\State End episode \textbf{if} $o_{t+1}$ is terminal.
\EndFor
\State Store episode in $\mathcal{D}$.
\EndWhile
\end{algorithmic}
\end{algorithm*}

\section{More Empirical Analysis}

\subsection{Prioritised Sampling \& Infrequent \textsc{look}}\label{baseline_ablation}
Our algorithm uses prioritised sampling and executes a \textsc{look} action every $I_{look}=20$ steps. The baseline agent LSTM-DRQN follows this algorithm. We now ask, 
\begin{quote}
    Does prioritised sampling and an infrequent \textsc{look} play a significant role in the baseline's performance?
\end{quote}
\begin{figure}
    \centering
    \includegraphics[width=\linewidth]{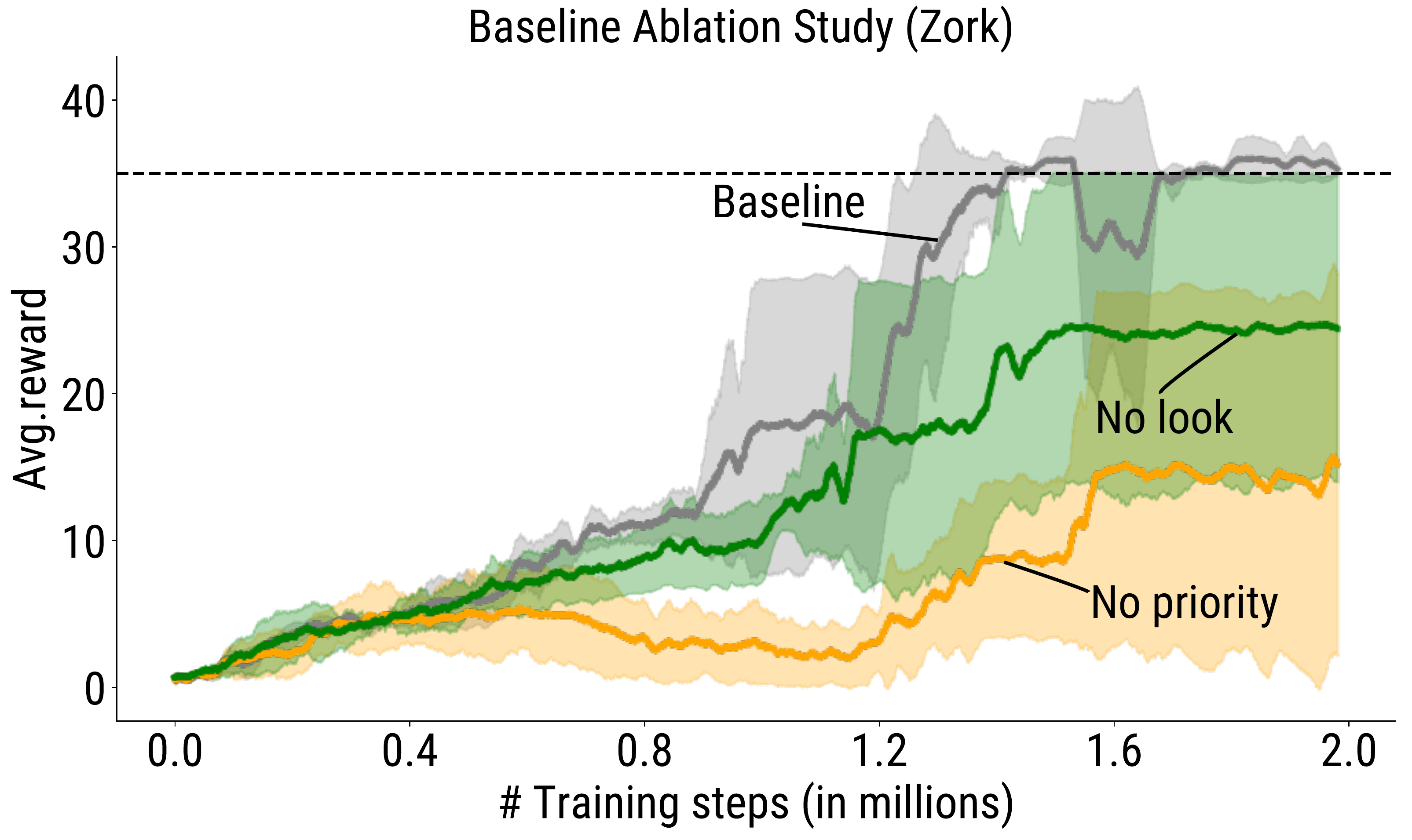}
    \caption{Learning curves for baseline ablation study.}
    \label{fig:zork_baseline}
\end{figure}
For this experiment, we compare the Baseline to two agents. The first agent is the Baseline without prioritised sampling and the second is the one without an infrequent look. Accordingly, we call them  ``No-priority (NP)" and ``No-\textsc{look} (NL)" respectively. We use Zork as the testing domain. 

From Fig \ref{fig:zork_baseline}, we observe that the Baseline performs better than the  NP agent. This is because prioritised sampling helps the baseline agent to choose the episodes in which rewards are received in, thus assigning credit to the relevant states faster and overall better learning. In the same figure, the Baseline performs slightly better than the NL agent. We hypothesise that even though \textsc{look} command is executed infrequently, it helps the agent in exploration and do credit assignment better.

\subsection{CQLH}\label{sec:alternate_cqlh}
Our algorithm uses CQLH implementation as described in Section \ref{action_gating_implementation}. An important case that CQLH considers is $s_{t+1}=s_{t}$. This manifests in $(1-\hatxi)$ term in equation (\ref{eqn:consistent_q_learning_histories}). We now ask whether ignoring the case $s_{t+1}=s_{t}$ worsen the agent's performance?
\begin{figure}
    \centering
    \includegraphics[width=\linewidth]{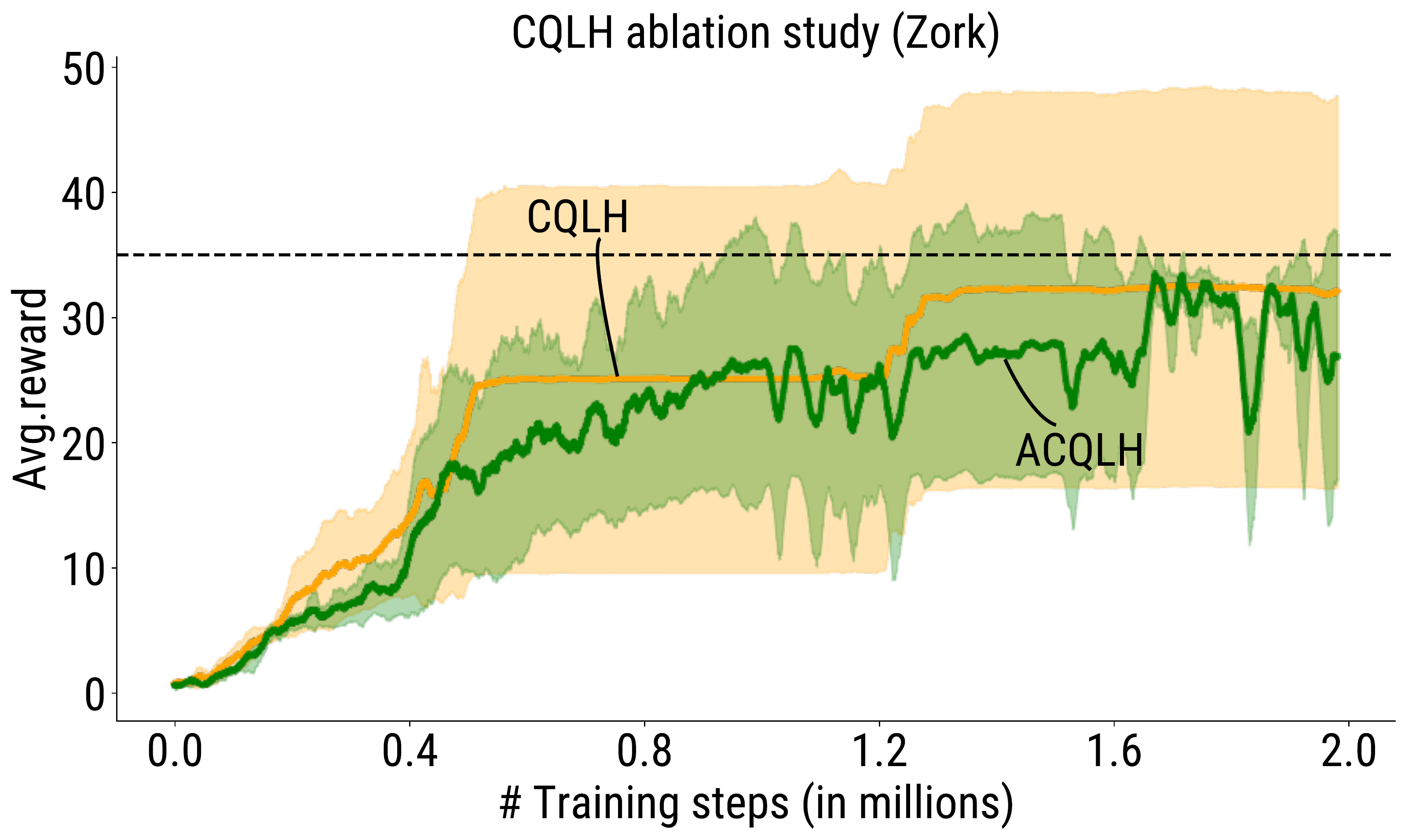}
    \caption{Learning curves for CQLH ablation study.}
    \label{fig:zork_cqlh}
\end{figure}
For this experiment, we compare CQLH agent with the agent which uses this error for update:
\begin{align*}
    \delta''_t = & r_t + \gamma \max_{a \in \cA} Q(h_{t+1}, a) \hatxi_t  - Q(h_t, a_t) .
\end{align*}
Accordingly, we call this new agent as ``alternate CQLH (ACQLH)'' agent. We use Zork as testing domain. From Fig \ref{fig:zork_cqlh}, we observe that although ACQLH has a simpler update rule, its performance seems more unstable compared to the CQLH agent.

\begin{figure*}
    \centering
        \begin{minipage}{.45\textwidth}
            \begin{subfigure}{\textwidth}
            \centering
            \includegraphics[width=\textwidth]{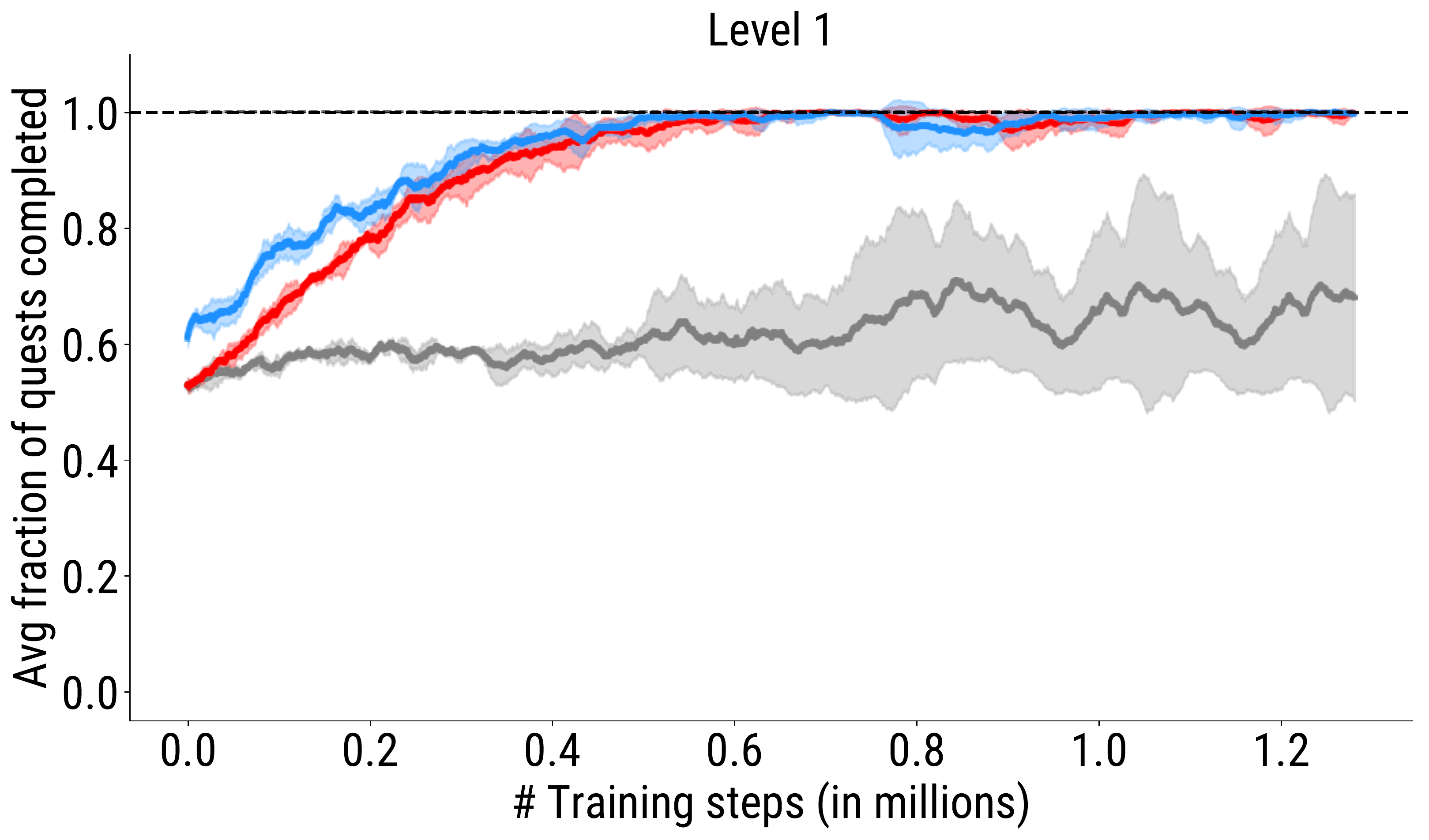}
    
            \end{subfigure}\\
            \begin{subfigure}{\textwidth}
            \centering
            \includegraphics[width=\textwidth]{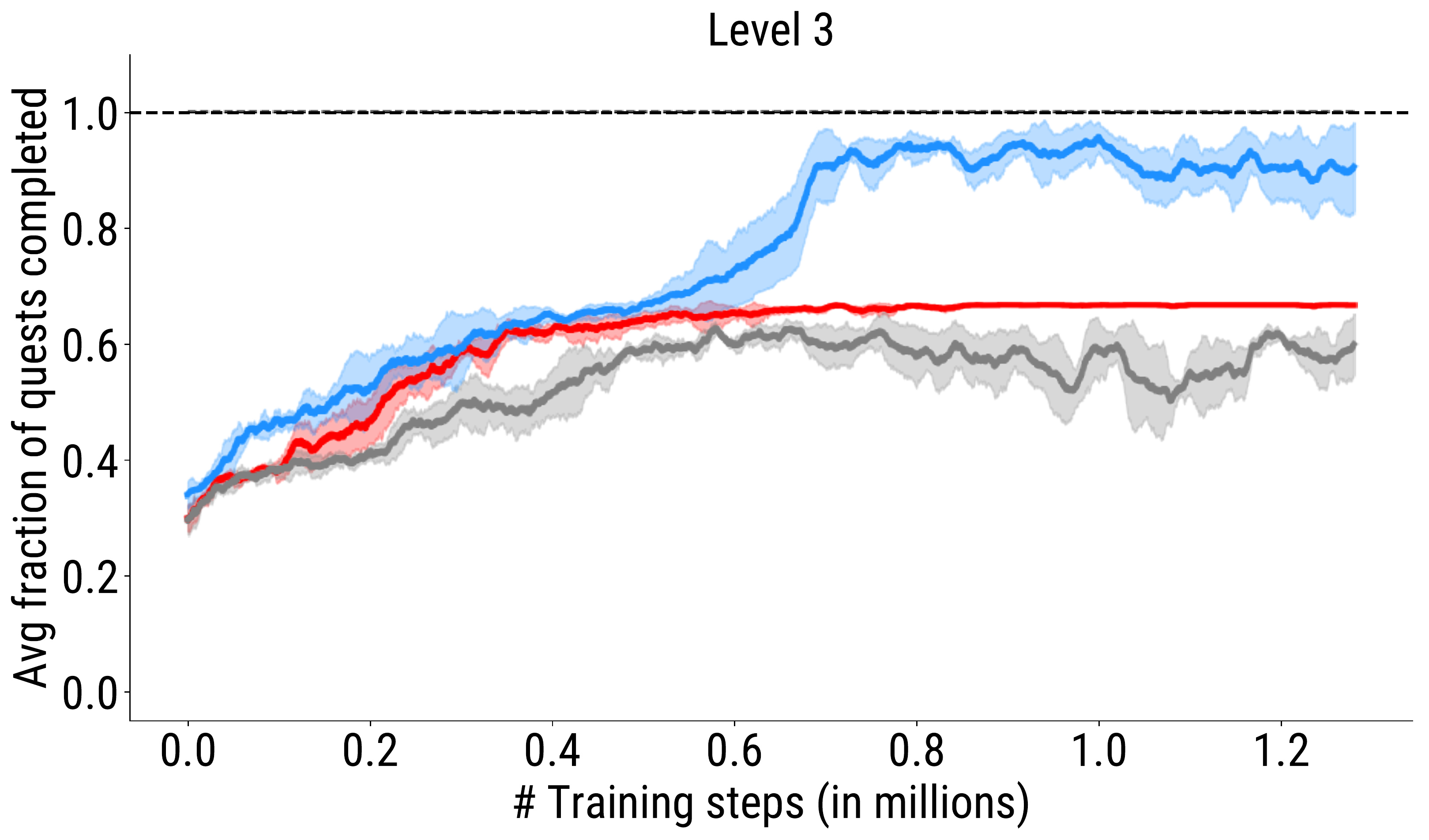}
            
            \end{subfigure}%
            
            \begin{subfigure}{\textwidth}
            \centering
            \includegraphics[width=\textwidth]{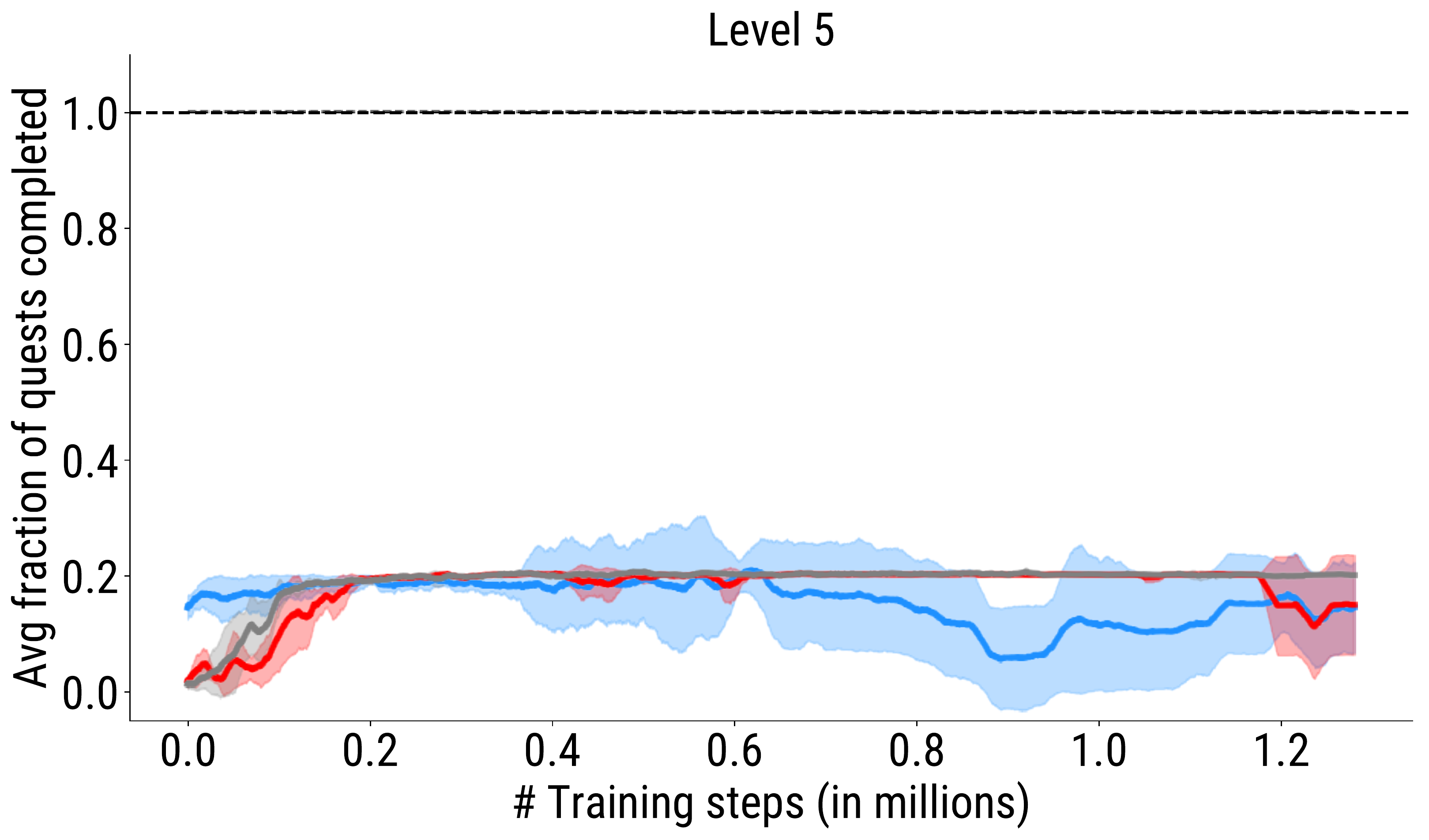}
            
            \end{subfigure}%
            
            \begin{subfigure}{\textwidth}
            \centering
            \includegraphics[width=\textwidth]{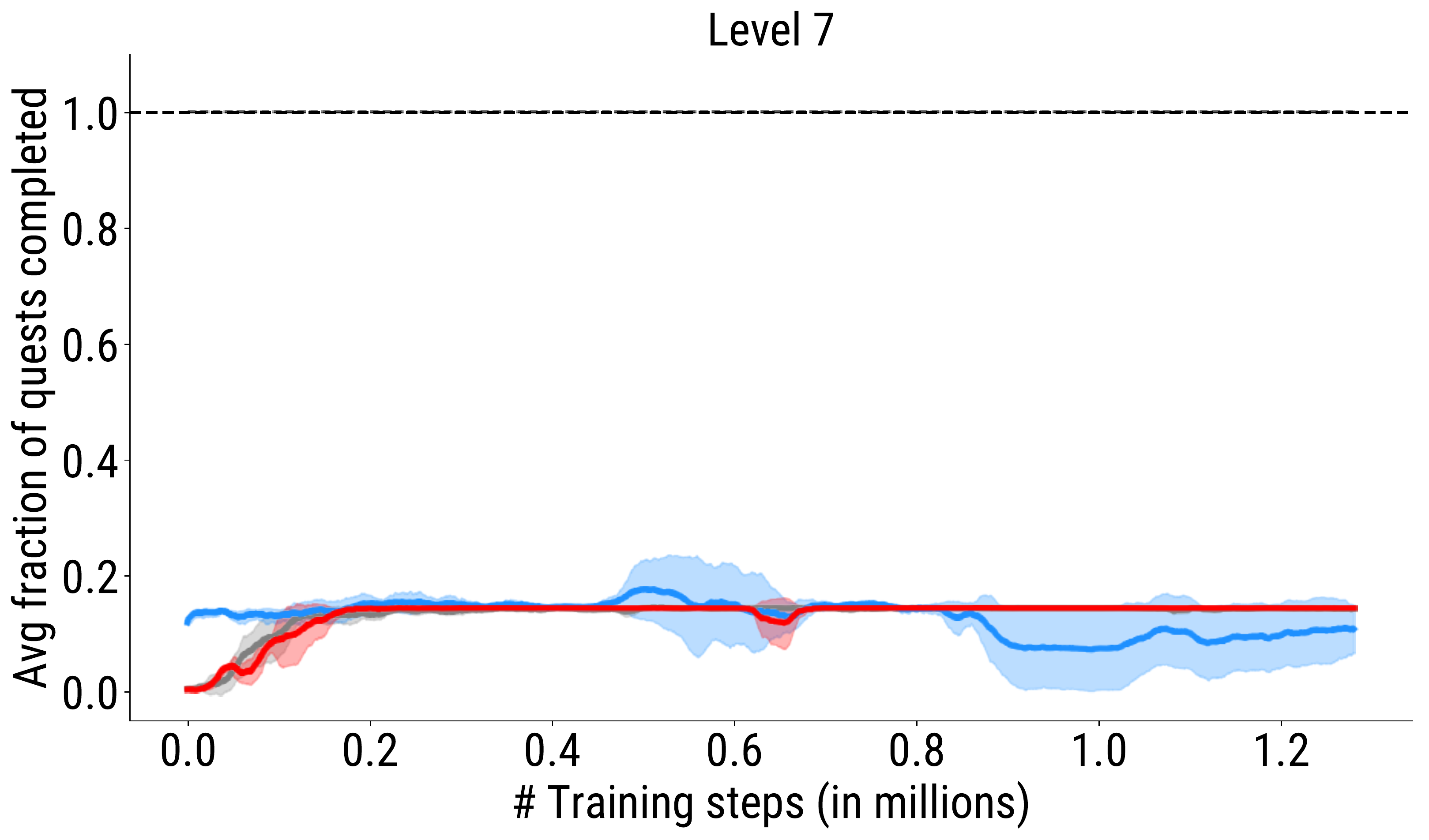}
            
            \end{subfigure}%
          
        \end{minipage}
        \hfill
        \begin{minipage}{.45\textwidth}
            \begin{subfigure}{\textwidth}
            \centering
            \includegraphics[width=\textwidth]{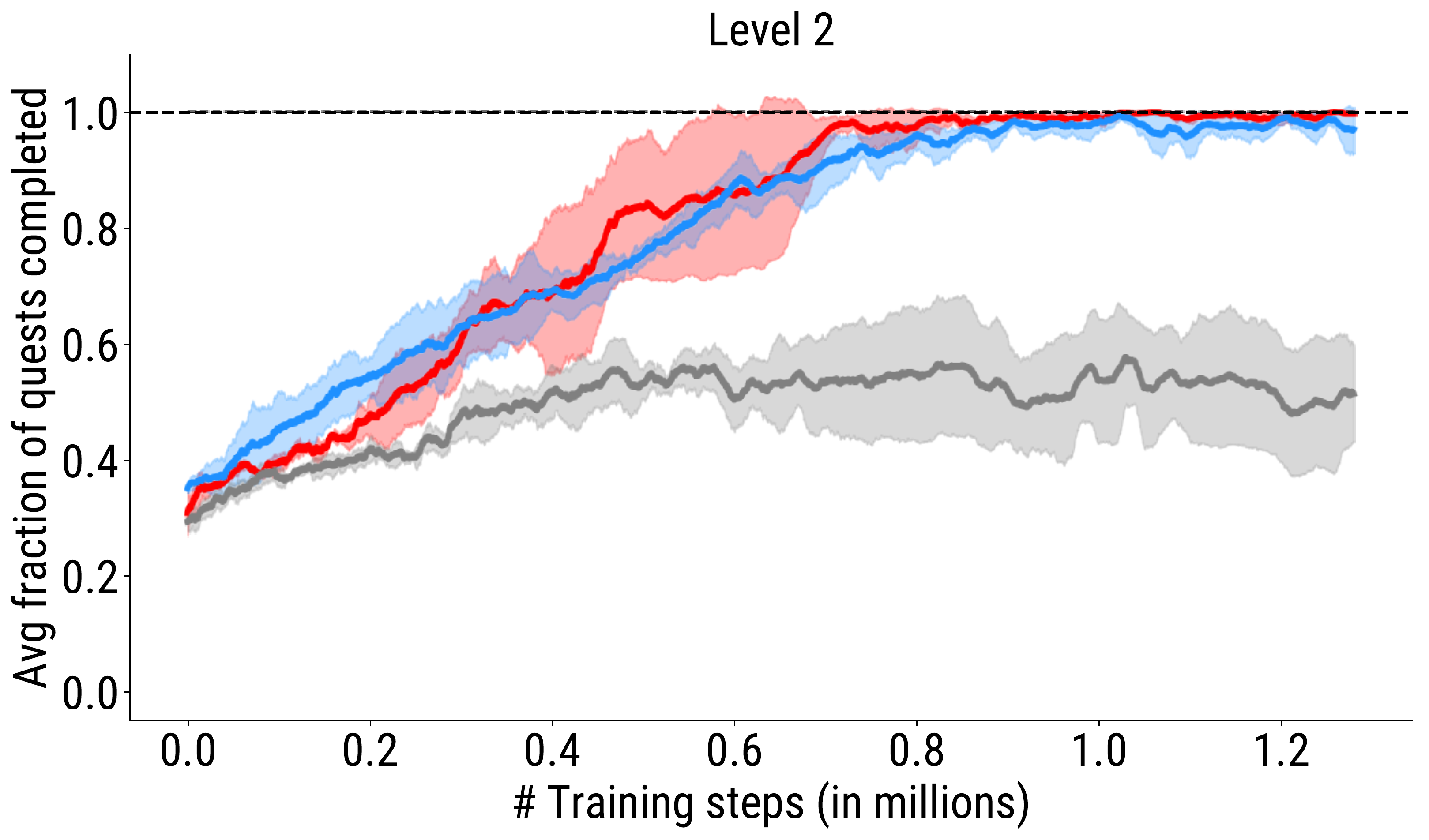}
    
            \end{subfigure}\\
            \begin{subfigure}{\textwidth}
            \centering
            \includegraphics[width=\textwidth]{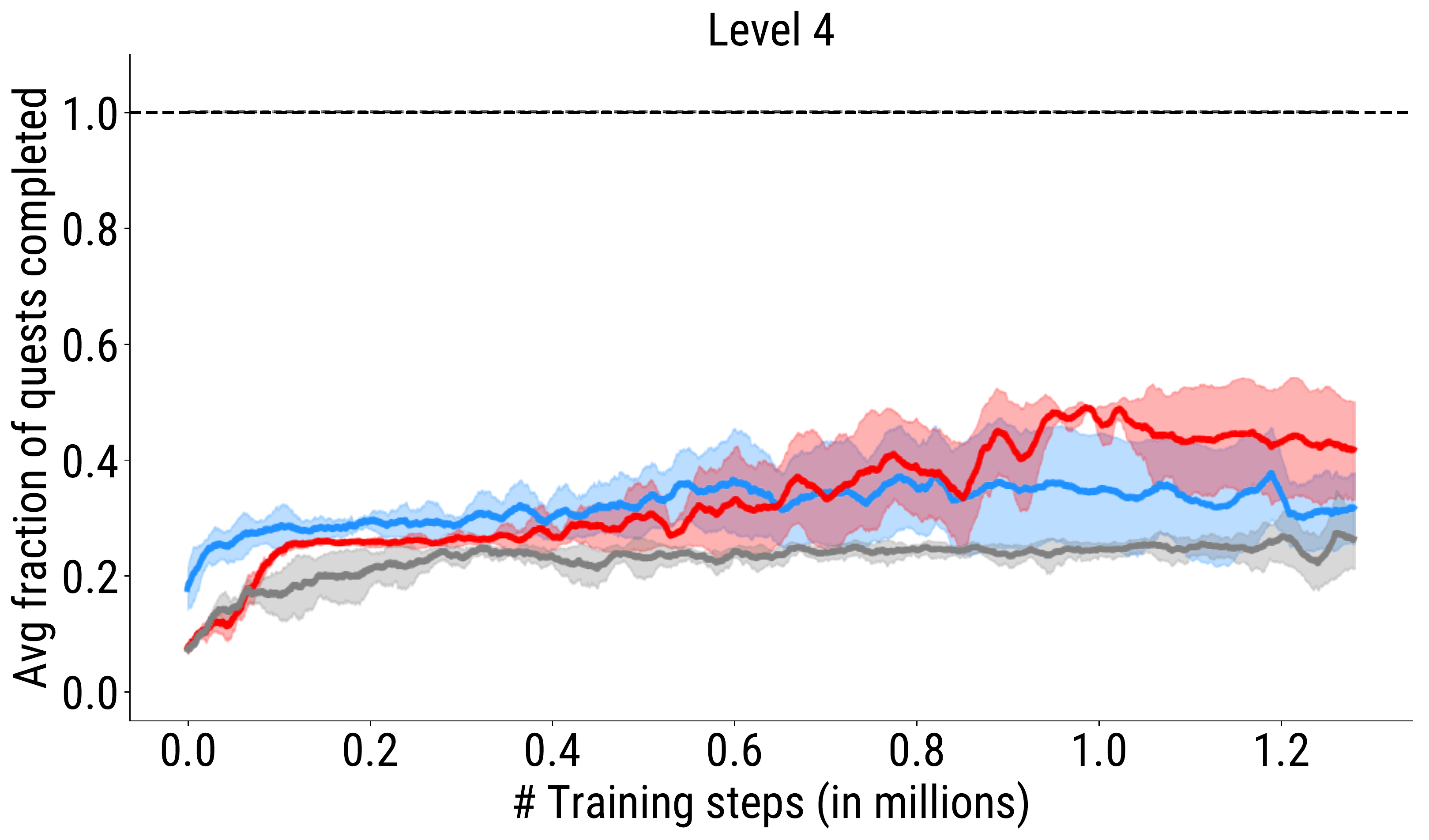}
            
            \end{subfigure}%
            
            \begin{subfigure}{\textwidth}
            \centering
            \includegraphics[width=\textwidth]{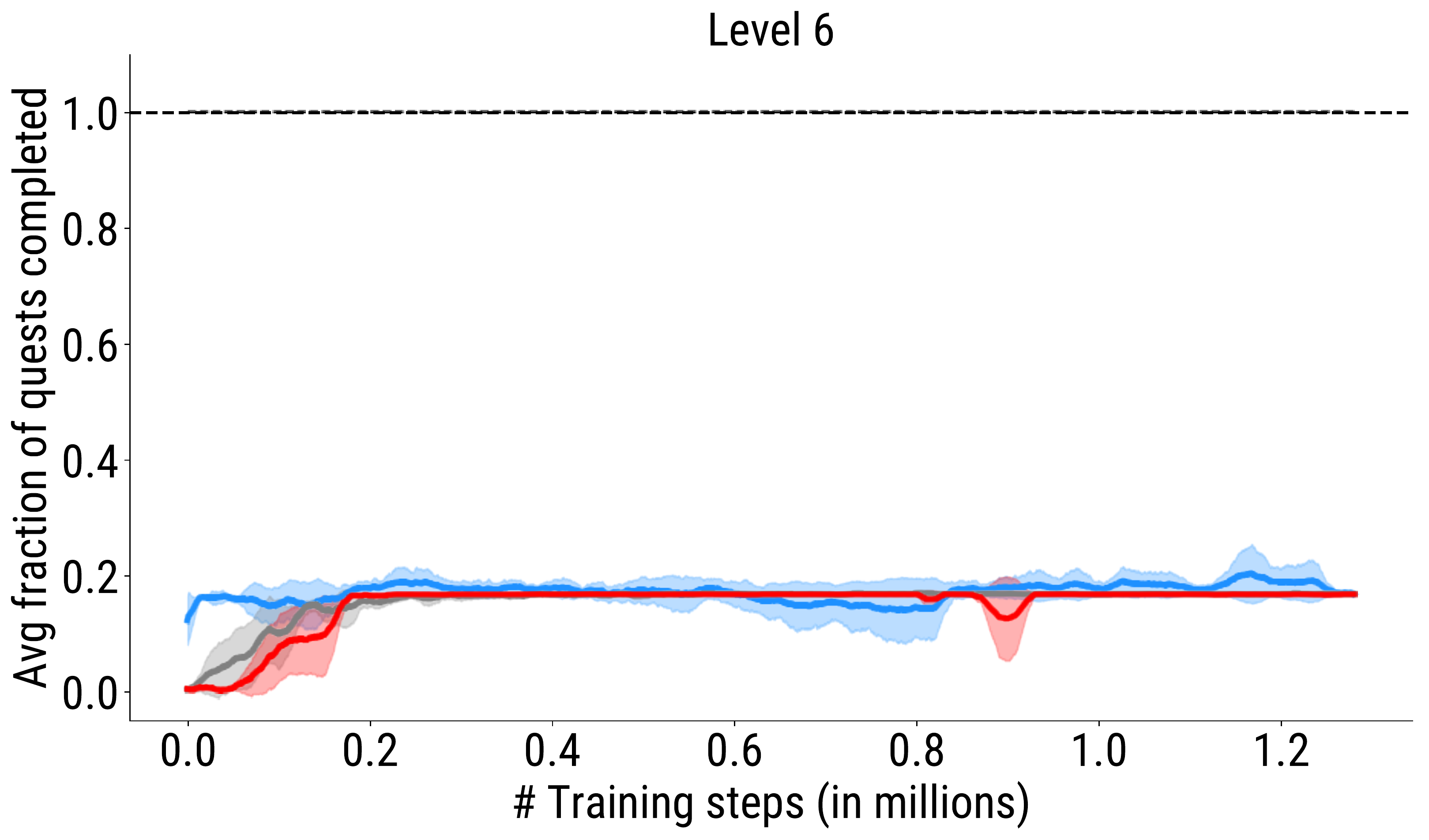}
            
            \end{subfigure}%
        
        \end{minipage}
        \hfill
\caption{Learning curves for Score contextualisation (SC : \textbf{red}), Score contextualisation + Masking (SC + Masking : \textbf{blue}) and Baseline (\textbf{grey}) for all the levels of the SaladWorld. For the simpler levels i.e. level 1 and 2, SC and SC + Masking perform better than  Baseline. With difficult level 3, only SC + Masking solves the game. For levels 4 and beyond, we posit that better exploration strategies are required.\label{fig:all_learning_curves_final_comparison}}
    \end{figure*}

\begin{figure*}
    \centering
    \includegraphics[width=\textwidth]{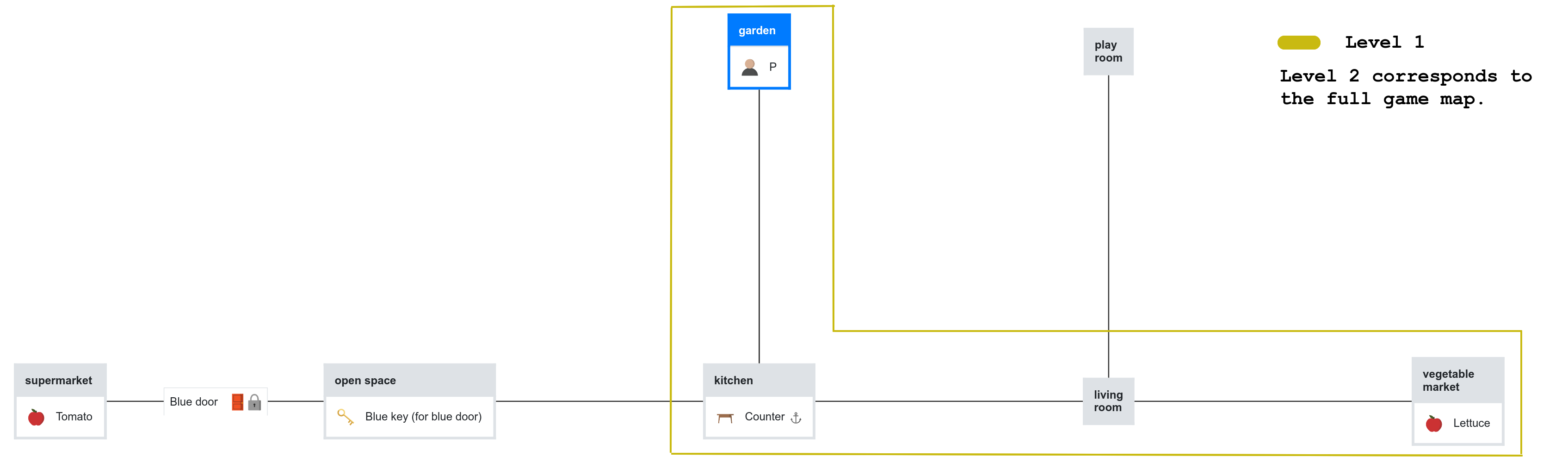}
    \caption{Game map shows Level 1 and 2.  Simpler levels help us test the effectiveness of score contextualisation architecture. }
    \label{fig:game map_level12}
\end{figure*}

\begin{figure*}
    \centering
    \includegraphics[width=\textwidth]{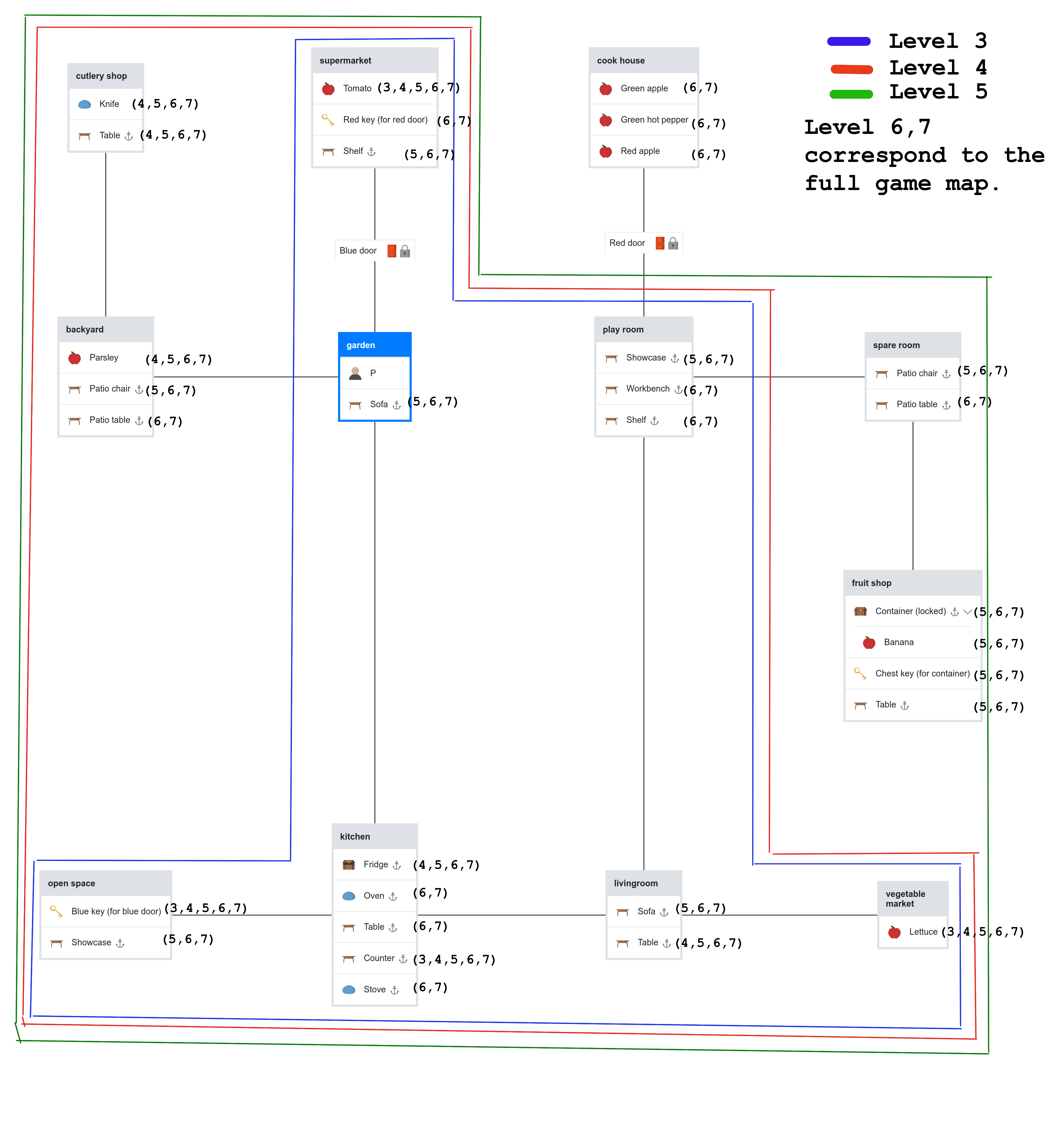}
    \caption{Game map shows the \emph{progression} from Level 3 to Level 7 in terms of number of rooms and objects. Besides every object in the map, there is a tuple which shows in which levels the object is available. We observe that with successive levels, the number of rooms and objects increase making it more difficult to solve these levels.}
    \label{fig:game map}
\end{figure*}

\end{document}